\documentclass[10pt,journal,compsoc]{IEEEtran}

\usepackage{microtype}

\usepackage{booktabs}

\usepackage{textcomp}
\usepackage{xcolor}
\usepackage{tcolorbox}
\usepackage{colortbl}
\usepackage{bm}

\usepackage{url}            %
\usepackage{nicefrac}       %
\usepackage{balance}
\usepackage{hyperref}
\usepackage{subcaption}

\usepackage{xspace}
\usepackage{color}
\usepackage{makecell}

\usepackage{amsfonts}
\usepackage{amssymb}
\usepackage{algorithm}
\usepackage{algpseudocode}
\usepackage{algorithmicx}
\usepackage{adjustbox}

\usepackage{xspace}
\usepackage{cleveref}
\usepackage{wrapfig}
\usepackage{enumitem}

\usepackage{framed}
\newenvironment{result}{\begin{framed}\centering\it}{\end{framed}}

\usepackage{multirow}
    
\usepackage{siunitx}
\usepackage{pifont}%

\let\emptyset\varnothing

\algnewcommand{\LineComment}[1]{\State\(\triangleright\) #1}

\newtheorem{definition}{Definition}

\newcommand{\approach}{\textsc{DistroFair}\xspace} 
\newcommand{\rev}[1]{\textcolor{black}{#1}}

\newcommand{\revise}[1]{\textcolor{black}{#1}}
\newcommand{\reviseTwo}[1]{\textcolor{black}{#1}}
\newcommand{\reviseThree}[1]{\textcolor{black}{#1}}

\newcommand{\recheck}[1]{\textcolor{black}{#1}}
\newcommand{\reviseCheck}[1]{\textcolor{black}{#1}}
\newcommand{\revision}[1]{\textcolor{black}{#1}}

\usepackage{bbding}

\usepackage[numbers]{natbib}

\def\tsc#1{\csdef{#1}{\textsc{\lowercase{#1}}\xspace}}
\tsc{WGM}
\tsc{QE}

\begin{document}

\title{Distribution-aware Fairness Test Generation}

\author{Sai~Sathiesh~Rajan, Ezekiel~Soremekun, Yves~Le~Traon, Sudipta~Chattopadhyay
\IEEEcompsocitemizethanks{\IEEEcompsocthanksitem S.S. Rajan, and S. Chattopadhyay are with Singapore University of Technology and Design. E. Soremekun is with Royal Holloway University of London. Y.L. Traon is with SnT, University of Luxembourg.}
}

\IEEEtitleabstractindextext{%

\begin{abstract}
Ensuring that all classes of objects are detected with equal accuracy is essential in AI systems. For instance, being unable to identify any one class of objects could have fatal consequences in autonomous driving systems. 
Hence, ensuring the reliability of image recognition systems is crucial.
This work addresses  
{\em how to validate group fairness in image recognition software}. 
We propose a \textit{distribution-aware fairness testing} approach (called \approach) that systematically exposes 
class-level fairness violations in image classifiers via a
synergistic combination of \textit{out-of-distribution (OOD) testing} and \textit{\reviseTwo{semantic-preserving} image mutation}. 
\approach automatically \textit{learns the distribution} (e.g., number/orientation) of objects in a set of images. 
Then it \textit{systematically mutates objects in the images} to become OOD 
using three \textit{\reviseTwo{semantic-preserving} image mutations} -- \textit{object deletion}, \textit{object insertion} and \textit{object rotation}. 
\reviseTwo{We evaluate \approach using
two well-known datasets (CityScapes and MS-COCO) and three major, \reviseTwo{commercial}
image recognition software (namely, Amazon Rekognition, Google Cloud Vision and Azure Computer Vision). Results show} 
that about 21\% of images generated by \approach \reviseTwo{reveal} class-level 
fairness violations \reviseTwo{using either ground truth or metamorphic oracles}. 
\approach is up to 2.3x more effective than \reviseTwo{two main \textit{baselines}, i.e., (a) an approach which focuses on generating images only \textit{within the distribution} (ID) and (b) fairness analysis using only the original image dataset}.  We further observed that \approach is efficient, it generates 
\reviseTwo{
460 
images per hour, on average.}
Finally, we evaluate the semantic validity of our approach via a user study with 81 participants, using %
30 real images and 30 corresponding mutated images generated by \approach. 
We found that images generated by \approach are 80\%
as realistic %
as real-world images. 

\end{abstract}

}

\maketitle

\section{Introduction}
\label{sec:intro}

Image classification has several critical applications in autonomous driving, robotics and healthcare, among 
others. Image classification may involve several tasks~\cite{wu2020multi}. For instance, given an image, one of the crucial tasks 
for several autonomous applications is to recognize the different objects in the image i.e., multi-label object 
classification (MLC)~\cite{wu2020multi}. 
Consider the MLC system used in autonomous driving, 
it is pertinent 
for the classifier to detect the objects on roads, including vehicles, pedestrians and  animals; all with {\em fairly} high 
accuracy. Failure to do so may lead to severe consequences, resulting in accidents. Indeed, 
image classification software have shown significant biases towards certain {\em class(es)}, e.g., dark-skinned people were 
more likely to be misclassified~\cite{racist-camera} and women were usually associated with activities such 
as cooking, shopping etc~\cite{men-like-shopping}. %
Disparities between class-level 
accuracy of a given image classification task may have several societal, legal and safety concerns. Therefore, 
systematic testing of image classification task, to detect potential bias against certain classes, is of critical 
importance. 

In this paper, we study the fairness of class-level accuracy in image classification tasks, specifically in MLC 
tasks. We choose MLC due to its applicability in several safety critical, autonomous applications e.g., driving 
and robotics. Given an arbitrary MLC model (\textit{system under test} ($\mathit{SUT}$)) and a set of 
initial images,  
our fairness test generation approach (called \approach) highlights the classes that face \textit{unusually high 
error rates}  for the $\mathit{SUT}$ to reveal an unfair treatment of one class as compared to others.
Additionally, each error 
is associated with concrete test images that can be used by the developer to further investigate the errors.

\revise{Our approach \reviseTwo{employs} 
 \textit{out-of-distribution} (OOD) testing. By learning the distribution of objects 
detected in an initial set of images, \approach systematically generates a set of images  \reviseTwo{that portrays a \textit{distributional shift} in the image dataset, such that the generated images are ``outside'' the learned distribution of objects in the initial sample. The generated images are called \textit{OOD images}}.
The \textit{key insight} behind our approach is to \textit{ensure that the fairness 
properties of an MLC system generalize to unlikely, yet possible scenarios via 
\reviseTwo{OOD} 
images}. 
We hypothesize that developers may ascertain fairness properties on likely scenarios (aka in-distribution) but 
ignore the unlikely scenarios, i.e., OOD.} 
\revise{For instance, consider a scenario where we generate a crowded road scene e.g., by inserting many pedestrians 
in an image that contained only a few pedestrian objects. \reviseTwo{Suppose we find that the accuracy of the 
``traffic light" class in such an OOD image is significantly 
lower than the accuracy of the ``car" class.  Then,} this implies that
the prediction of the ``traffic light" class is 
\reviseTwo{\textit{unfair} in comparison to}
the ``car" class. Such a different treatment 
for the two classes 
\reviseTwo{violates} 
statistical parity~\cite{verma2018fairness}.
\approach works both in the presence and absence of ground truth, making it general and applicable also 
to unlabelled/partially labeled datasets.}
\revise{{\em To the best of our knowledge, we present the first OOD testing approach to discover and analyze the 
class-level fairness errors in image classification tasks.}}

\begin{figure}
	\centering
	\includegraphics[scale=0.32]{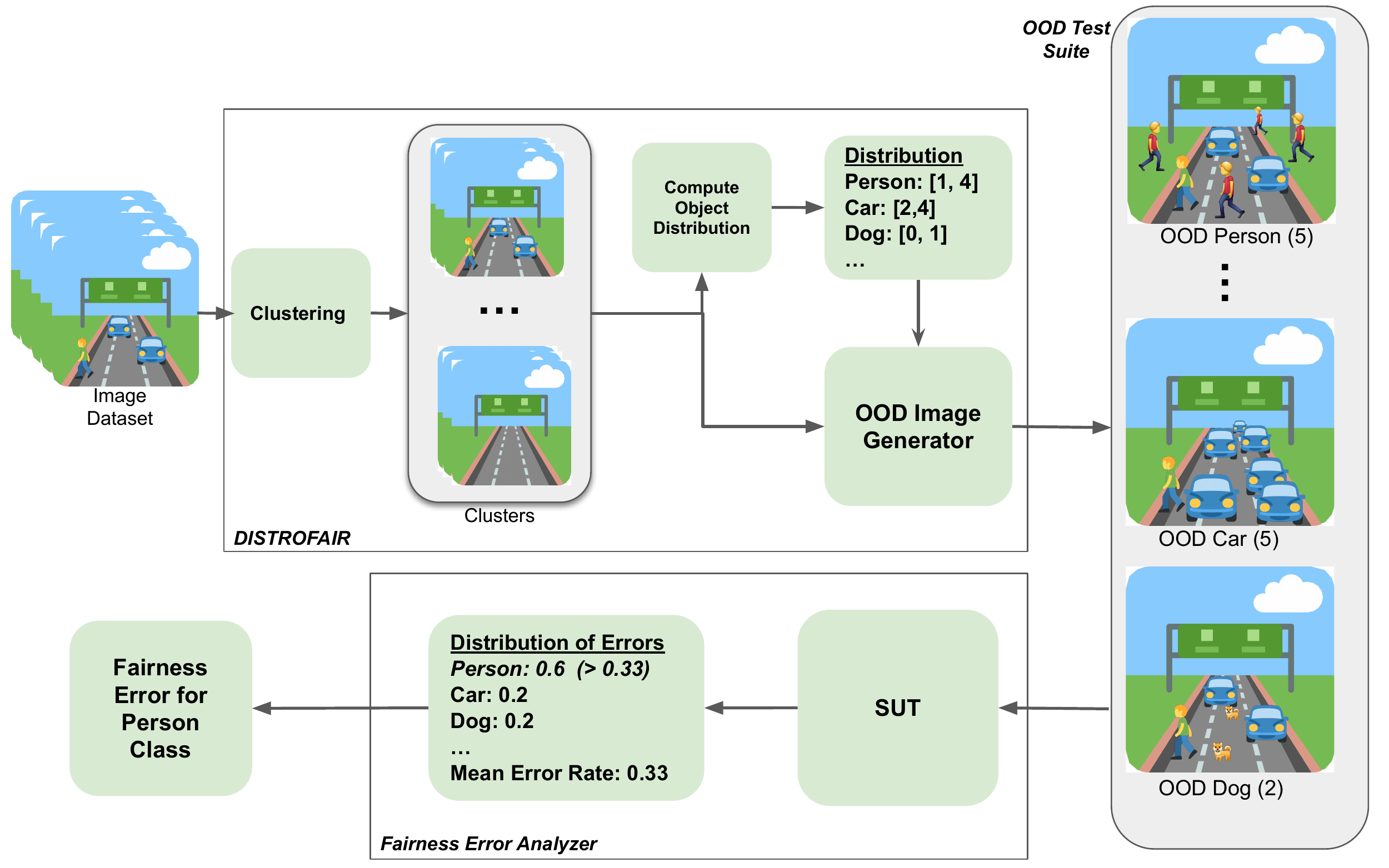}
	\caption{
      An illustration of our  \approach approach.
	}
	\label{fig:approach-steps}
\end{figure} 

\autoref{fig:approach-steps} illustrates the different steps of our \approach approach. 
\approach  starts with randomly sampled images from a dataset. This sample set of 
images are then clustered into different similar sub-groups to take into account the 
diversity of images in the initial sample. For each sub-group/cluster, \approach then 
computes a distribution of objects detected by the $\mathit{SUT}$. Such distribution includes 
information about the minimum and maximum number of objects detected for each class and their 
orientation. Subsequently, OOD images are generated by leveraging this information and using 
semantic-preserving mutation operators (e.g., insertion, deletion and rotation of objects). For instance, 
three OOD images are shown in \autoref{fig:approach-steps}, each one exceeds the maximum 
number of ``{\em Person}'', ``{\em Car}'' or ``{\em Dog}'' objects detected by the $\mathit{SUT}$ in the respective 
cluster. Finally, the $\mathit{SUT}$ is subject to analysis on the generated OOD images. 
As observed in \autoref{fig:approach-steps}, if a class (e.g., ``{\em Person}'') is detected with an error rate 
(i.e., $0.6$) more than the mean error rate across all classes (i.e., $0.33$),  then \approach highlights the 
class (i.e., {\em ``person}'') as facing a fairness error. Although we target our evaluation for MLC tasks, our OOD testing  approach 
is general and can be applied to other multi-label image classification tasks.

Despite several approaches on fairness testing~\cite{udeshi2018automated,zhang2020white,soremekun2022astraea} and 
functional testing~\cite{deepexplore,deeptest} of machine-learning based systems, systematic fairness testing 
of class-level errors is relatively less explored. Our approach is complementary to recent effort in 
detecting class-level confusion and bias errors in deep learning models~\cite{tian2020testing}. In particular, 
while the aforementioned work presented new metrics for confusion and bias detection for a 
class~\cite{tian2020testing}, we propose \revise{an} OOD test generation approach to complement the detection 
of class-level fairness errors. Recent works on image fuzzing are focused on generating semantically valid 
images~\cite{semantic-image-fuzzing-icse2022} or detecting functional errors without evaluating semantic 
validity~\cite{metamorphic-image-ase2020}~\cite{image_captioning_issta2022}. In contrast, we propose a novel OOD test generation method for  
systematically discovering class-level fairness errors. We also evaluate the semantic validity 
of generated OOD images via a user study. 

This paper makes the following contributions:
\begin{enumerate}
\item We formalize how to measure class-level fairness errors \reviseCheck{in image recognition software} and propose a novel OOD test generation approach (\approach) to discover 
such errors 
(\autoref{sec:methodology}). 

\item We 
\reviseTwo{propose and implement} three metamorphic OOD transformations such that the resulting images 
are semantically valid with high likelihood (\autoref{sec:methodology}). 

\item Based on the OOD images, we propose an automated approach to detect the class-level fairness errors in image classification 
tasks (\autoref{sec:methodology}). 

\item %
We implement our \approach approach and evaluate it with three image 
classification systems from major vendors (Google, Amazon and Microsoft) using two datasets (MS-COCO and CityScapes). 
Our evaluation generates $\approx$24K error-inducing OOD images (out of a total $\approx$ 112K  OOD images), finding nearly 
\revise{368} classes (\revise{out of a total 879 classes}) facing fairness errors across different models, datasets, OOD style 
mutations and \revise{fairness test oracles} (\autoref{sec:results}). 

\item We compare our OOD test generation approach with \reviseTwo{two main baselines, namely (a) fairness analysis using \textit{only} the original dataset, and (b)} a test generation approach tailored to generating inputs 
\textit{within distribution} (ID). We show that our OOD test generation approach improves the discovery of fairness error rate by 
up to 131.48\% (\autoref{sec:results}).  

\item We \reviseTwo{conduct} %
a user study to evaluate the semantic validity of our OOD images. Our study reveals that our generated 
OOD images are about 
80\% as realistic as 
original, real-world images, on average (\autoref{sec:results}). 

\end{enumerate}

We discuss threats to validity (\autoref{sec:threats}). We then describe closely related work (\autoref{sec:related-work}) 
before concluding (\autoref{sec:conclusion}).

\section{Overview}
\label{sec:overview}

\begin{figure}
	\centering
	\includegraphics[scale=0.32]{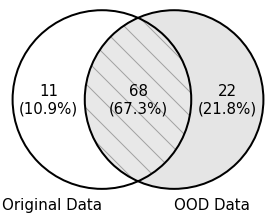}
	\caption{Classes with higher than the mean error rate in original sample vs. OOD sample generated from the original.
	}
	\label{fig:venn-Org-OOD}
\end{figure}

\begin{table*}
\centering
\caption{Outline of \approach: \textcolor{blue}{Inclusion errors [Inc.]} are highlighted in \textcolor{blue}{blue}, \textcolor{red}{exclusion errors [Ex.]}  are highlighted in \textcolor{red}{red}, and \underline{GT errors} are \underline{underlined}. The numbers within (parenthesis) in column 3 and column 5 capture respective ground truths.}
\resizebox{\linewidth}{!}{ 
\label{overviewTable}
\begin{tabular}{ccccc}
\toprule
\begin{tabular}{@{}c@{}}Subject/ \\ Mutation\end{tabular}
& Original Image & 
\begin{tabular}{@{}c@{}}Detected \\ Objects\end{tabular}
& Mutated Image &
\begin{tabular}{@{}c@{}}Detected \\ Objects\end{tabular} \\
\midrule
\begin{tabular}{@{}c@{}}MS (Insertion) \\ Cat\end{tabular}
&
\adjustbox{valign=c}{\includegraphics[scale=0.2]{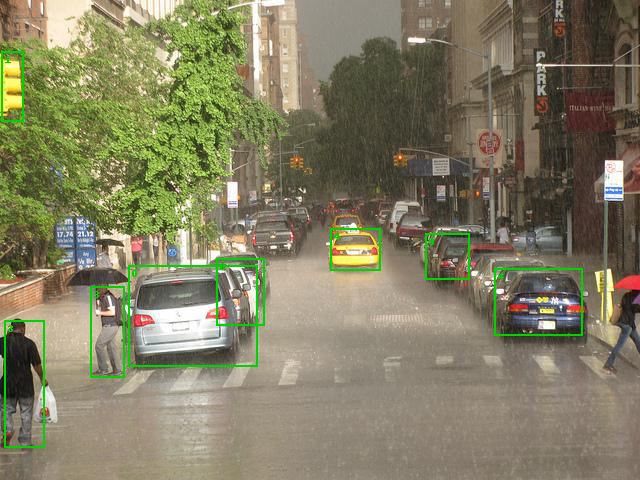}}
& \begin{tabular}{@{}c@{}}Car: 3 (15)\\ Person: 2 (7)\\ Taxi: 2 (2) \\ Traffic Light: 1 (5)\end{tabular}
&
\adjustbox{valign=c}{\includegraphics[scale=0.2]{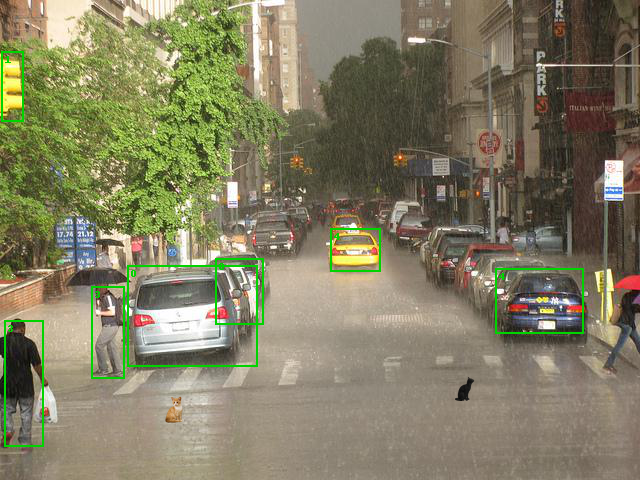}}
& \begin{tabular}{@{}c@{}} \textcolor{red}{[Ex.] \underline{Car}: 2 (15) } \\ Person: 2 (7)\\ Taxi: 2 (2)\\ Traffic Light: 1 (5)\end{tabular} \\
\midrule
\begin{tabular}{@{}c@{}}AWS (Deletion) \\ Person\end{tabular}
&
\adjustbox{valign=c}{\includegraphics[scale=0.2]{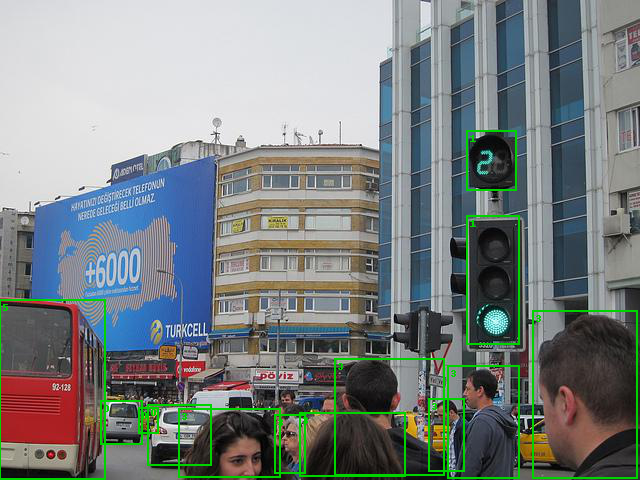}}
& \begin{tabular}{@{}c@{}}Car: 3 (8)\\ Person: 7 (12)\\ Traffic Light: 2 (3)\\ Bus: 1(1)\end{tabular}
&
\adjustbox{valign=c}{\includegraphics[scale=0.2]{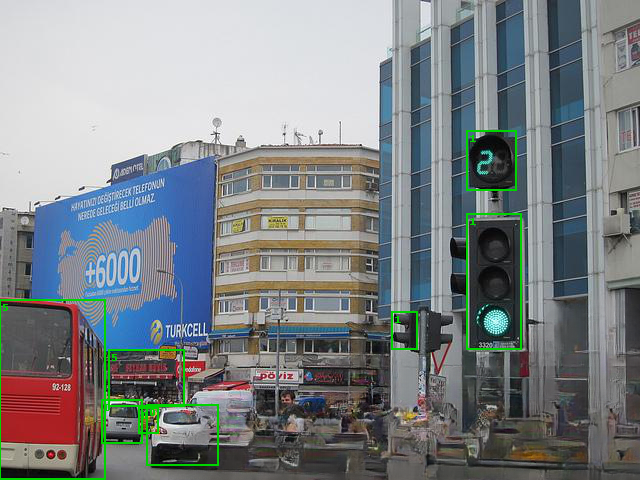}}
& \begin{tabular}{@{}c@{}}Car: 3 (8)\\ \textcolor{blue}{ [Inc.] Traffic Light: 3 (3)} \\ \textcolor{blue}{[Inc.] \underline{Bus}: 2 (1) }\end{tabular} \\
\midrule
\begin{tabular}{@{}c@{}}GCP (Rotation) \\ Person\end{tabular}
&
\adjustbox{valign=c}{\includegraphics[scale=0.2]{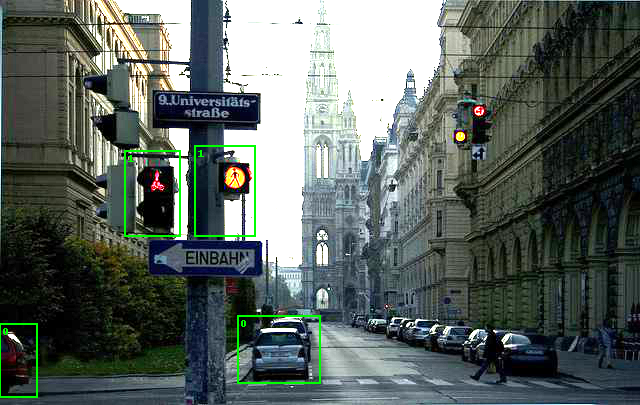}}
& \begin{tabular}{@{}c@{}}Car: 2 (12) \\ Traffic Light: 2 (6)\end{tabular}
&
\adjustbox{valign=c}{\includegraphics[scale=0.2]{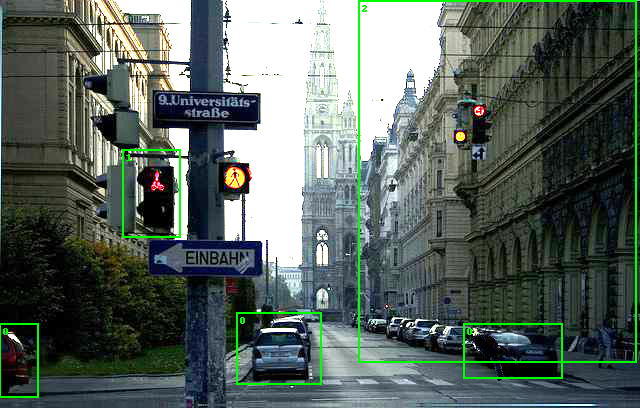}}
& \begin{tabular}{@{}c@{}} \textcolor{blue}{[Inc.]Car: 3 (12)} \\ \textcolor{red}{[Ex.]\underline{Traffic Light}: 1 (6)} \\ \textcolor{blue}{[Inc.]Building: 1 (1)}\end{tabular} \\
\bottomrule
\end{tabular}
\label{tab:approach-overview}
}
\end{table*}

In this section, we outline the motivation behind our approach and illustrate it with an example. 

\smallskip\noindent
\textbf{%
\reviseTwo{Class-level fairness}:}
In this work, we investigate and discover class-level fairness errors in computer vision (CV) systems. Class level fairness is directly 
related to the concept of group fairness. Fundamentally, group fairness is concerned with ensuring that different groups exhibit 
similar statistical properties under similar stimuli~\cite{verma2018fairness}. For instance, a CV system that recognizes different 
classes of objects (e.g., cars, people) with a similar degree of precision and recall is said to be fair. 
 \revise{This formulation is appropriate in safety-critical situations such as autonomous driving, where accurately identifying all objects 
 on the road is desirable. More concretely, an autonomous car with a CV system that preferentially detects vehicles when compared to 
 pets or animals is unfair. We note that such a formulation does not make any assumptions on the protected group(s). } 
Concretely, for two arbitrary class labels $a$ and $b$, we expect that class-level fairness is satisfied if and only 
if the following holds for a model $f$ with a set of classes $\mathbb{C}$: 
\begin{equation}
\label{eq:abstract-class-fairness}	
Pr(f(a)) \approxeq Pr(f(b))\  \  \  \forall a, b \in \mathbb{C}	
\end{equation}	
where $Pr(f(a))$ and $Pr(f(b))$ capture the probability that class $a$ and class $b$ are correctly classified by $f$, respectively. 

\smallskip\noindent
\textbf{\reviseTwo{Key Insight (Why OOD samples?)}:}
OOD testing is increasingly becoming popular to evaluate the capability of an ML-based system beyond the training 
set~\cite{vqa-cp, ood-bench}. 
In particular, significant manual effort has been put forward to create OOD benchmark~\cite{OOD-model}. Moreover, we have observed 
a line of research that focused on improving the accuracy of ML models on OOD benchmark~\cite{OOD-engineering}. 
\revision{This is particularly important in autonomous driving systems since the training data is unlikely to capture the full set of scenarios that can occur in the real world. In addition, certain scenarios might not be easily obtained from real world testing on account of it being dangerous or time consuming. For instance, the likelihood of encountering a car that is being driven on the wrong side of the road is comparatively low and the number of such scenarios captured in the training set is likely to be small. As such, we postulate that OOD generation approach that seeks to generate these unseen scenarios is likely to be effective at exposing weaknesses in image recognition systems, including fairness issues.}
 
In this paper, we 
propose a methodology to automatically generate OOD images from arbitrary image samples
to validate %
class-level fairness of %
a target ML model. %
\revise{Our \textit{key insight} is driven by the observation of \textit{distributional shift} in class-level accuracy \reviseTwo{between an original dataset and their corresponding OOD images}. \autoref{fig:venn-Org-OOD} illustrates the set of classes that have higher than the mean accuracy 
across three widely used object recognition models from Microsoft, Amazon (AWS), and Google (GCP). This 
is shown both for a sample of original data (taken from an existing dataset) and the OOD images created from 
this sample using \approach. Concretely, we observe that the accuracy of 21.8\% of classes drops below 
the mean accuracy {\em only when considering the OOD images}. From this observation, \textit{we posit that 
inducing distributional shifts} (such as those illustrated in \autoref{fig:venn-Org-OOD}) \textit{may unmask hidden biases}. Therefore, it is 
desirable to investigate class-level biases \reviseTwo{
in the OOD dataset w.r.t. to its distributional shift from the original dataset}.
} 
Our generation of OOD images considers scenarios that may occur in real world. 
Thus, the class-level accuracy on the 
OOD images provides the model developers useful debugging information. For instance, such information may highlight 
the specific classes where the model performs poorly when stressed with generated OOD images.

\smallskip\noindent
\textbf{An illustrative example:}
 \autoref{tab:approach-overview} shows an example illustrating our  OOD-image generation and the class-level error detection. 
 All the illustrated errors are taken from our evaluation on real-world system from Microsoft (MS), Amazon (AWS) and Google (GCP). 
 The first column shows the targeted subject (MS/AWS/GCP) and the mutation operation (e.g., insertion, deletion, rotation of object). 
 The second column captures the original image and the third column highlights the class-level detection on 
 the original image by the respective subject. The fourth column captures the OOD image based on the mutation shown in the 
 first column and the rightmost column captures the subject output on the mutated images. 
 
 Intuitively, given a dataset $S$ and a model $M$, we capture the distribution of any class $c \in \mathbb{C}$ 
 ($\mathbb{C}$ being the set of all classes) as follows: we record the minimum and maximum occurrences of 
 class $c$ detected by $M$ for any image $s \in \mathbb{S}$. Additionally, we also record the 
orientation (angle)  
 in a similar fashion for all classes. The generation of OOD images for model $M$ thus focuses on creating an image 
 that deviates from the captured distribution. For instance, consider the insertion operation in 
 \autoref{tab:approach-overview} for MS. In our evaluation, we observed that MS did not detect any {\em cat} 
 class for our original sample set. Thus, we consider the insertion of even a single {\em cat} object will result in 
 an OOD image. In the example shown in \autoref{tab:approach-overview}, we insert two {\em cat} objects as 
 shown in the mutated image. As a consequence, MS fails to detect one of the {\em car} objects that was detected 
 in the original image. 
 \revise{In general, we consider two different test oracles as follows to detect errors in the generated OOD images:}

 \begin{enumerate}[leftmargin=*]
	\item {\em Ground Truth} ({\bf GT}) based Oracle: \revise{A class $c$ in an OOD image faces error if and only if 
	the detection accuracy of $c$ with respect to the ground truth drops below the detection accuracy of $c$ in the 
	corresponding original image. For example, the insertion operation shown in \autoref{tab:approach-overview} drops 
	the detection accuracy of the {\em Car} class in the OOD image (from $\frac{3}{15}$ to $\frac{2}{15}$). Hence, one 
	error is accounted for the {\em Car} class. In contrast, the detection accuracy of {\em Traffic Light} class improves with the 
	deletion operation for AWS. As the detection accuracy improves with respect to ground truth, we do not count 
	such phenomenon as an error. 
	Nonetheless, we also account for such improvement in accuracy, as our approach is targeted to compute 
	fairness metrics across classes. Hence, our approach allows for negative errors to consider cases where 
	the detection of a class improves with mutation.  
	Formally, \reviseTwo{the} number of errors for an unmodified class $c$ (via the mutation operation) 
	is accounted as follows:}
\begin{equation}
\label{eq:err-count}
	Err_c = \left |  num_{ood}(c) - GT_{c} 	\right |  -  \left |  num_{orig}(c) - GT_{c} 	\right |
\end{equation}
\revise{where 	$num_{ood}(c)$, $num_{orig}(c)$ and $GT_{c} $ capture the number of class $c$ objects detected in 
the OOD image, in the corresponding original image and the ground truth for class $c$ in the original image, 
respectively.}

	\item {\em Metamorphic} ({\bf MT}) Oracle: \revise{It is often infeasible in practice to use the ground truth data due to 
	the unavailability of perfectly labeled data. Moreover, class detection varies across subjects, 
	tasks and contexts. For example, a speed camera detects only license plates, whereas surveillance systems track multiple 
	objects. Similarly, even for the same class, models might prioritize foreground objects over background objects. Consequently, 
	a universal ground truth may not capture the intent of the model under test. To address this, we also design a metamorphic 
	(MT) oracle that considers changes in detection accuracy with respect to the detection accuracy in the original image. 
	In other words, we capture the intent of the targeted model in line with its accuracy in the original, unmodified image. 
    Then, we investigate whether the prediction of different classes \reviseTwo{are  consistent %
with respect to 
OOD style mutations. 
}} 

\vspace{0.1cm}
	
	\revise{Concretely, we consider errors in two categories: {\em (i)} {\em Inclusion error} means that some object from 
	a given class was {\em not detected} in the original image, but it is detected in the corresponding OOD image. 
	{\em (i)} {\em Exclusion error} means that some object from a given class was detected in the original 
	image, but it is {\em not detected} in the corresponding OOD image. As illustrated in \autoref{tab:approach-overview}, 
	the deletion operation leads to one inclusion error for the classes {\em Traffic Light} and {\em Bus} in AWS. 
	On the contrary, the insertion operation results in one exclusion error in MS for the {\em Car} class, whereas the rotation 
	operation leads to an exclusion error in GCP  for the {\em Traffic Light} class.}
	\revise{We compare the effectiveness of both the GT and MT  oracles in {\bf RQ1}.} 
	
\end{enumerate}

{\em We exclude any errors due to the 
mutated class} (i.e., the {\em cat} class for insertion operation). This is to eliminate the potential impact of bias in our 
experiments, as the mutated class is often likely to have more errors than the unmodified classes.

Our OOD image mutation is carefully engineered to generate 
semantically valid images. For example, while inserting an object, \approach tries to compute the appropriate size of the 
respective object in the image. 
This is accomplished by heuristically estimating the size of the inserted object with respect to the size of existing 
objects in the image. %
 For example, as observed from \autoref{tab:approach-overview}, 
our mutation inserts appropriately sized {\em cat} objects. Likewise, the other mutations keep the classes in the 
OOD image recognizable.

\smallskip\noindent
\textbf{Computing fairness errors:}
Starting with a dataset $S$, we apply all the operations (insertion/deletion/rotation) to get the set of OOD images $S'$. 
For a given model $M$, we then compute the number of exclusion and inclusion errors for each class 
$c \in \mathbb{C}$ over the dataset $S'$. Such errors provide an overall distribution of errors across all classes in the 
OOD image set. We consider that a class $c \in \mathbb{C}$ exhibits fairness errors when its error rate exceeds the 
mean error rate across all classes. For example, if $\mathit{Err}_{c}$ captures the error rate for class $c$, then a class 
$c'$ exhibits fairness error if and only if $\mathit{Err}_{c'} > \frac{\sum_{c \in \mathbb{C}} \mathit{Err}_c}{|\mathbb{C}|}$. 
We note that $\mathit{Err}_{c}$ is computed as the ratio between the total number of errors faced by class $c$ in $S'$ 
and the total number of objects of class $c$, in dataset $S$. 
In \autoref{tab:approach-overview}, \revise{using the MT oracle}, the 
{\em car} class has an error rate of 33\% (=1/3) for MS considering just one image in $S'$. Likewise for AWS, the classes {\em Traffic Light} 
and {\em Bus} have error rates of 50\% and 100\%, respectively.

\section{Methodology}
\label{sec:methodology}

In this section, we first formally define the notion of OOD images 
considered within \approach. Then we discuss \approach in detail.  
\approach can broadly be considered to have three components, 
namely, clustering, an OOD image generator and a fairness error analyzer. 
In the following, we elaborate each of the three components.

\begin{definition}
{\em (OOD Image)} Let us assume an initial set of images $\mathit{Img}_{list}$ where the number of objects of a given class $c$ 
is bounded by $\langle min_c, max_c \rangle$. Additionally, the set of orientations (angles) for any object of class $c$ within $\mathit{Img}_{list}$ 
is captured by $\Theta_{c}$. We call an image $\mathcal{I}'$ an OOD image with respect to the initial set of 
images $\mathit{Img}_{list}$ and the given class $c$ if and only if one of the following conditions hold: 
{\em (i)} $\mathcal{I}'_c < min_c$  {\em (ii)} $\mathcal{I}'_c > max_c$  {\em (iii)} $\Theta(\mathcal{I}'_c) \notin \Theta_c$. 
$\mathcal{I}'_c$ captures the number of objects of class $c$ in image $\mathcal{I}'$ whereas $\Theta(\mathcal{I}'_c)$ captures the 
set of orientations of class $c$ objects in $\mathcal{I}'$.
\end{definition}

\subsection{Clustering}
\label{sec:image-clustering}
Our \approach approach starts with an arbitrary sample of images. We first employ clustering on the initial sample to create smaller groups 
of images. This grouping is performed for images with similar objects and scenery. Additionally, the clustering handles variance of images/objects 
and the mixture of distributions in our initial sample. Specifically, our \approach approach determines, for each image in the initial sample, the 
number of objects for each class. We leverage \revise{a} state-of-the-art detection and segmentation library i.e., Detectron2~\cite{wu2019detectron2} 
for this purpose. The class-level information for all images is then fed to a clustering algorithm to divide the initial sample into similar subgroups. 
In general, our approach can leverage any clustering algorithm. We use K-Means clustering algorithm~\cite{kmeans} within \approach. Once the 
clusters of images are computed, OOD image generation is employed on each cluster of images independently. In the following, we discuss OOD 
image generation for an arbitrary cluster of images.

\begin{algorithm}[t]
	{\scriptsize
		\begin{algorithmic}[1] 
			\Procedure{OOD\_Image\_Generation}
			{$\mathit{Img}_{List}, \mathit{OP}_{List}, \mathit{LBL}_{List}$}
			\State $OOD\_Set \gets \emptyset$
			\State $\mathit{MUT}_{List} \gets \{\mathit{x}, \mathit{y}\} \colon \mathit{x} \in (\mathit{OP}_{List}), \mathit{y} \in (\mathit{LBL}_{List})$
			\LineComment $\mathcal{F}$ computes the distribution of the set of images in $\mathit{Img}_{List}$  
			\LineComment $\mathit{SUT}$ is the ML model under test
			\State $\mathit{Dist}_{List} \gets \mathcal{F} (\mathit{Img}_{List}, \mathit{SUT})$
			
			\For{$\mathit{M} \in \mathit{MUT}_{List}$}
			\For{$\mathit{Img} \in \mathit{Img}_{List}$}				
			\State $\mathit{Dist}_{Img} \gets \mathit{F}_{Img} (\mathit{Img}, \mathit{SUT})$
			\State $\mathit{Mut}_{Num} \gets \mathit{MutGen}(\mathit{Dist}_{Img}, \mathit{Dist}_{List})$
			\State $\mathit{Gen}_{Img} \gets \mathit{ImageGen}(\mathit{Img}, \mathit{M}, \mathit{Mut}_{Num})$
			\State $OOD\_Set \,  \boldsymbol{\cup \!=} \{(\mathit{Gen}_{Img}, \mathit{Img}, \mathit{M})\}$			
			\EndFor
			\EndFor
			\State \Return $OOD\_Set$
			\EndProcedure
	\end{algorithmic}}
	\caption{OOD Image Generation.}
	\label{alg:test-gen}
\end{algorithm}

\subsection{OOD Image Generation}
\label{sec:ood-image-generation}
Algorithm~\ref{alg:test-gen} outlines our OOD image generation process for a target ML model $\mathit{SUT}$.
\reviseCheck{In the beginning, \approach learns a distribution, $\mathit{Dist}_{List}$, for the set of images $\mathit{Img}_{List}$ under test.} The knowledge of this distribution is leveraged for OOD image generation process. 
Concretely, for each class $c \in \mathbb{C}$, the distribution captures a 
triplet $\langle \Theta_c, \mathit{min}_c, \mathit{max}_c \rangle$. $\Theta_c$ captures the set 
of %
orientations (angles) %
for objects in class $c$ and $\mathit{min}_c$ (respectively, $\mathit{max}_c$) captures 
the minimum (respectively, maximum) number of objects of class $c$ detected by the $\mathit{SUT}$ 
in $\mathit{Img}_{List}$. 
After computing the distribution $\mathit{Dist}_{List}$, \approach aims to generate OOD images for each 
image in the $\mathit{Img}_{List}$. To this end, we consider a list of mutation operators $\mathit{MUT}_{List}$ 
where each $M \in \mathit{MUT}_{List}$ is a pair, containing the operation (insertion/deletion/rotation) and the 
target class for mutation. 
For generating an OOD image, \approach identifies the distribution for a single image, $\mathit{Dist}_{Img}$. 
$\mathit{Dist}_{Img}$ is used to compute the exact characteristics of the mutation for an OOD transformation. 
For example, given $\mathit{Dist}_{Img}$ and $\mathit{Dist}_{List}$, we compute the possible number insertions 
(e.g., $\mathit{MUT}_{num}$ in Algorithm~\ref{alg:test-gen}) of class $c$ objects such that the total number of 
class $c$ objects exceeds $max_c$. This is then used to produce the OOD image $\mathit{Gen}_{Img}$ via 
the procedure $\mathit{ImageGen}$. All successfully generated OOD images are stored for subsequent analysis 
of class-level fairness errors.

\subsection{Mutation Operators}\label{sec:mutation-operators}

\smallskip \noindent
\textbf{\reviseTwo{Semantic-preserving Mutations}:} \reviseTwo{In this work, mutation operators are designed to preserve the image semantics i.e., \textit{the meaning of the image 
in the real world}~\cite{semantic-image-fuzzing-icse2022}. 
The goal %
is to preserve the perception of the original image, except for 
the mutated object(s). 
Our mutation operators rely on state-of-the-art tools for fine-grained image 
modifications. However, due to the current limitations of these tools, there is no guarantee that the semantics are 
always preserved in the OOD images. To mitigate this, we conducted a user study ({\bf RQ4}) to check the semantic 
validity of generated images.} 
\revise{In the following, we discuss the design details of the three mutation operators (\textit{see}~\autoref{tab:approach-overview}).}

\smallskip \noindent
\textbf{\textit{Insertion}:} The insertion operation of \approach employs several heuristics to ensure semantic correctness.  We describe and illustrate this operation using \autoref{tab:insert-table}.

We first determine the relative size of each object to be inserted by comparing it to a reference object. 
As an example,  
let us consider a ``car'' 
to be the reference object.  We first find relative heights, w.r.t.  the number of pixels,  for all the other objects (e.g.,  ``person'' or ``bicycles'') to be inserted by determining how much larger or smaller they are 
in comparison to the reference object (``car'').  We determine these relative sizes via initial experimentation. 
This is a one time effort that is leveraged for all insertion operations going forward.

\approach then leverages a technique named panoptic segmentation~\cite{Kirillov_2019_CVPR}~\cite{wu2019detectron2} on the original image 
to find the class label of each pixel. This is used to determine the size and location of the object to be inserted. 
\begin{table*}
\centering
\caption{Table illustrating the different steps present in the insertion operation. First, the heights of the objects in the image are obtained and rate at which the size changes as the position changes is noted. We then determine whether the object can be safely placed on the ground in the chosen location and reject it in cases where it would not be on the ground. We then show the final result of the insertion operation.}
\begin{tabular}{cc}
\toprule
Original Image & Resizing Process \\
\midrule
\adjustbox{valign=c}{\includegraphics[scale=0.33]{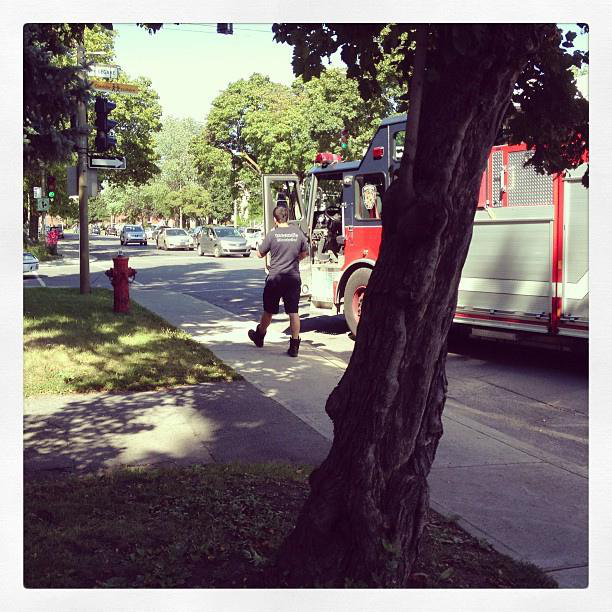}}
&
\adjustbox{valign=c}{\includegraphics[scale=0.33]{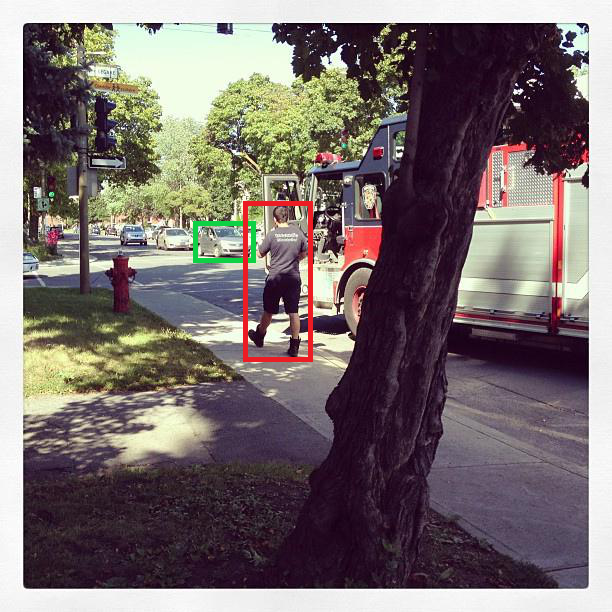}} \\
\midrule
Insertion Locations & Final Image \\
\midrule
\adjustbox{valign=c}{\includegraphics[scale=0.33]{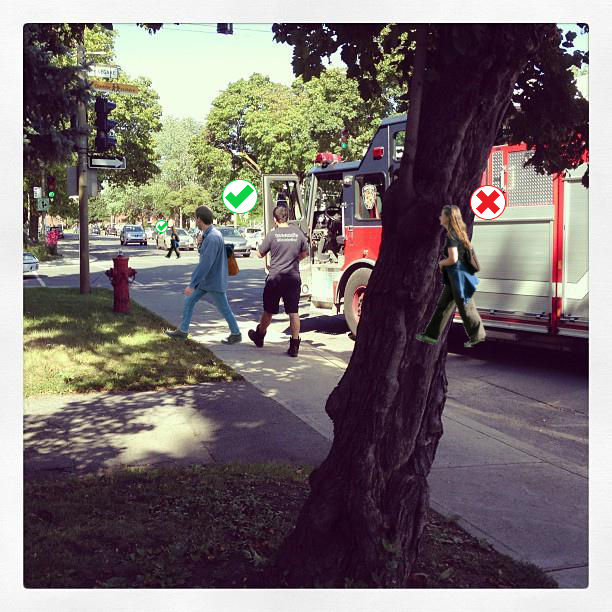}}
&
\adjustbox{valign=c}{\includegraphics[scale=0.33]{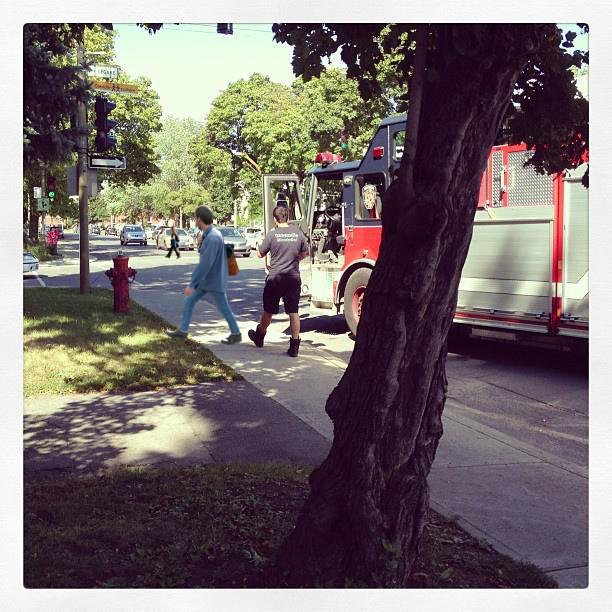}} \\
\bottomrule
\end{tabular}
\label{tab:insert-table}
\end{table*}
In particular, we aim to determine what size the reference object would be if it were to be inserted in the middle of the image. 
In addition, we also %
determine how the size of the reference object would vary if the object were moved one pixel up 
or down. For instance, consider the image in Table~\ref{tab:insert-table}.  We observe that there are multiple cars and a person 
present in the image. 
Let us consider the person in the red box (top right image in \autoref{tab:insert-table} (Resizing Process)). \approach uses our segmentation map to determine the height of the person in terms of the number of pixels. It then leverages our pre-computed relative reference sizes to determine the height of a hypothetical car in the same position.  Similarly, it measures the height of the car in the green box.  Next,  it combines 
these %
information to 
compute the scaling factor, i.e., how the height of the car changes with respect to the changes in y-coordinate of a two-dimensional image. 
The scaling factor is employed to find the size of the reference ``car'' if it were to be situated in the center of the image. By passing this information along with the scaling factor, \approach is able to easily determine the size of any object to be inserted in the original image, irrespective of its type or location with the aid of our knowledge of relative sizes.
We further note that computing the scaling factor for each original image is a one-time effort, i.e.,  it is not recomputed for  future insertion operations.

To identify appropriate locations for each insertion operation, \approach implements checks to ensure that the object is placed at an appropriate location. For instance, it ensures that a person is placed 
on the ground (road, pavement, or dirt) by checking whether the pixels that will be occupied by the bottom portion of the object are classified as belonging to the ground.  \autoref{tab:insert-table} (bottom left image) illustrates appropriate ({\color{green} \checkmark}) and inappropriate ({\color{red} $\times$}) locations for the insertion operation.  In addition, it ensures that objects that are further away from the perspective of the observer are placed before objects that are nearer to the observer.  This is to avoid 
objects in the background inadvertently obscuring objects in the foreground. \revision{
Furthermore, \approach determines the feasibility of placing the requisite number of objects into the original image.  In the event that \approach is unable to place enough objects to generate an OOD image, 
\rev{the insertion operation is not performed on the image.
\approach skips the image %
and attempts mutating 
the next image in the dataset.} This prevents us from forcibly inserting objects into a crowded image that might be unable to accommodate the additional objects.
}

\smallskip \noindent
\textbf{\textit{Deletion}:} %
During deletion, \approach deletes all object instances that belong to the class being mutated. 
We note that such deletion operation is an extreme case of OOD mutation when object deletion is considered 
for a given class. We choose this option to keep our test generation simple.  \approach leverages the panoptic segmentation 
map to identify the objects before applying a mask. It then uses inpainting~\cite{bertalmio2000image}~\cite{suvorov2022resolution} 
to delete the masked objects.

\smallskip \noindent
\textbf{\textit{Rotation}:} %
For  rotation, \approach first identifies an object belonging to the target class, taking care to ensure that said object is not 
obstructed by another object. It then extracts the image level information for the object being rotated in that location before 
deleting the object through inpainting~\cite{bertalmio2000image}~\cite{suvorov2022resolution}. Finally, it rotates the 
extracted image (i.e., the target object) and inserts it back into the original image. During insertion, we ensure that the physical 
dimensions such as height and width remain unchanged for the rotated object.

\begin{algorithm}[t]
{\scriptsize
		\begin{algorithmic}[1] 
			\Procedure{Fairness\_Error\_Counter}{$OOD\_Set, Case\_Type, Err\_Type$}
			
			\State $Tot\_Count \gets \emptyset$
			\State $Err\_Count \gets \emptyset$
			
			\For{$TUP \in OOD\_Set$}
			
			\LineComment {number and type of objects in the image as found by $\mathit{ORACLE}$ and $\mathit{SUT}$}
			\State $Oracle \gets \mathit{ORACLE} (TUP, Case\_Type, Error_Type)$
			
			\State Let $Oracle = (GT, Diff)$

			\For $(-, Obj_L) \in Diff$
			
				\LineComment{Accumulate the total number of objects and errors for each class} \label{ln:error-start}
				\State $Tot\_Count[Obj_L] \, + \! = GT[Obj_L]$
				\State $Err\_Count[Obj_L] \, + \! = Diff[Obj_L]$ \label{ln:error-end}
				
				\LineComment{Increment error count for class $Obj_L$}
				\If {$Diff[Obj_L] > 0$}
				\State $Err\_ImgDict[Obj_L] = Err\_ImgDict[Obj_L] + 1$    	
				\EndIf
			\EndFor
			
			\EndFor
			\EndProcedure
	\end{algorithmic}}
	\caption{Fairness Error Analysis.}
	\label{alg:fairness-analysis}
\end{algorithm}

\begin{algorithm}[t]
{\scriptsize
		\begin{algorithmic}[1] 
			\Procedure{ORACLE}{$TUP, Case\_Type, Err\_Type$}
			
			\If {$Case\_Type = ``SUT"$}
			
			\LineComment {number and type of objects in the image as found by $\mathit{SUT}$}
			\State $GT\_Pred \gets \mathit{SUT} (TUP.\mathit{Img})$
			
			\ElsIf {$Case\_Type = ``GT"$}
			
			\LineComment{$\mathit{GT}$ is the union of all SUT results and the dataset ground truth}
			\State $GT\_Pred \gets \mathit{GT} (TUP.\mathit{Img})$
			
			\EndIf
			
			\State $Org\_Pred \gets \mathit{SUT} (TUP.\mathit{Img})$ 
			\State $Gen\_Pred \gets \mathit{SUT} (TUP.\mathit{Gen}_{Img})$
			
			\State Let $\mathit{TUP.M} = (-, \mathit{LBL})$	\label{ln:init-start}

			\State $Diff\_Pred[\mathit{LBL}] \gets \emptyset$    
			\State $GT\_Pred[\mathit{LBL}] \gets \emptyset$
			\State $Org\_Pred[\mathit{LBL}] \gets \emptyset$
			\State $Gen\_Pred[\mathit{LBL}] \gets \emptyset$ \label{ln:init-end}

			\For{$(-, Obj_L) \in GT\_Pred$} \label{ln:compare-start}
			
				\State $Org\_Err \gets \left|Org\_Pred[\mathit{Obj_L}] - GT\_Pred[\mathit{Obj_L}]\right|$
				
				\State $New\_Err \gets Gen\_Pred[\mathit{Obj_L}] - GT\_Pred[\mathit{Obj_L}]$\label{ln:compare-end}
				
				\If {$Case\_Type = ``GT"$} \label{ln:error-identify-start}
				
				\State $Diff\_Pred[Obj_L] = \left|New\_Err\right| - Org\_Err$
				
				\ElsIf {$Case\_Type = ``SUT"$}
				
				\If {$Err\_Type = ``INC"$}
				
				\If {$Change\_Error > 0$}
				
				\State $Diff\_Pred[Obj_L] = New\_Err$
				
				\EndIf
					
				\ElsIf {$Err\_Type = ``EXC"$}
				
				\If {$Change\_Error < 0$}
				
				\State $Diff\_Pred[Obj_L] = -1 \cdot New\_Err$ \label{ln:error-identify-end}
				
				\EndIf
				
				\EndIf
				
				\EndIf	
				
			\EndFor
			
			\State $Oracle\_Ret = \{GT\_Pred, Diff\_Pred\}$
			
			\State \Return $Oracle\_Ret$
			
			\EndProcedure
	\end{algorithmic}}
	\caption{Fairness Error Oracle.}
	\label{alg:fairness-oracle}
\end{algorithm}

\subsection{Fairness Error Analysis}
Algorithm~\ref{alg:fairness-analysis} outlines our fairness error analysis. Given a set of OOD images (computed via Algorithm~\ref{alg:test-gen}), 
Algorithm~\ref{alg:fairness-analysis} computes, for each class, the total number of detected objects ($Tot\_Count$) and the number of errors in 
the detection ($Err\_Count$). 
\revise{To compute the number of errors, it relies on Algorithm~\ref{alg:fairness-oracle} to find the expected number of objects in each class. 
Algorithm~\ref{alg:fairness-oracle} takes the type of error being computed, and returns the appropriate number of errors for each class along with 
the initial reference.}

\revise{To obtain the ground truth reference i.e., $\mathit{GT}_{Ref}$ for each image, we use the following equation:
\begin{equation}
\label{eq:gt-reference}	
\mathit{GT}_{Ref}(Img)  = \mathit{GT}_{Data}(Img)\cup \bigcup_{i \in \mathit{All\_SUT}} \mathit{SUT}_{\mathit(i)}(Img)
\end{equation}
In essence, we take the reference to be the multiset union of the results from each subject under test  and the ground 
truth from the provided data ($\mathit{GT}_{Data}$).
We then set the expected count for the class of object being mutated to be zero, both in the original image and the corresponding OOD image 
(Line~\ref{ln:init-start}-Line~\ref{ln:init-end}).
}
This prevents us from inadvertently including errors that were directly introduced by the mutated objects themselves. 
Intuitively, in the absence of errors, we expect the original and corresponding OOD image to detect the same number 
of objects for each class, except the mutated class. We then find the degree to which the detected output for the mutated images has changed from the original output. (Line~\ref{ln:compare-start}-Line~\ref{ln:compare-end}). 
This is used to compute the errors (Line~\ref{ln:error-identify-start}-Line~\ref{ln:error-identify-end}). 
Algorithm~\ref{alg:fairness-analysis} then accumulates the error counts for all the images in the set of OOD 
images (Line~\ref{ln:error-start}-Line~\ref{ln:error-end}). It also calculates the number of images in which 
a particular class is exhibiting errors.

Once the errors for each class is computed in $Err\_Count$, we can compute the error rate for each class as follows: 
\begin{equation}
\label{eq:error-rate}	
\mathit{Err}_c  = \frac{Err\_Count[c]}{Tot\_Count[c]},\ \ \ \ \forall c \in \mathbb{C}
\end{equation}	
Finally, 
a class $c$ 
faces a
fairness error when its detection error rate exceeds the mean error rate across all classes:
\begin{equation}
	\label{eq:fairness-rate}	
	\mathit{Err}_c  > \frac{\sum_{i \in \mathbb{C}}Err_i}{|\mathbb{C}|}
\end{equation}	
In summary, the developer can use our framework to investigate the distribution of errors faced by each class and observe 
the classes exhibiting unusually high error rates. Additionally, %
each error is associated with a test case 
that allows the 
developer to investigate and reproduce the error.

\paragraph{\bf \em Usage of the OOD Tests:}
\revision{Given the set of classes \rev{that induce} 
fairness errors, developers can direct their efforts towards improving 
\rev{model performance for unfair classes. }This could be achieved in several ways. For instance, developers could direct their data collection teams to obtain more instances of the unfair classes to augment their training data. Developers could also augment the training set with the \rev{error-inducing} images generated by \approach. Recent research~\cite{deepexplore,deeptest} has shown that the addition of error-inducing inputs to the training data improves the accuracy of computer vision models. We further note that our technique inherently generates scenarios that are previously unseen in the initial dataset. As such, the addition of the error-inducing inputs to the training data could conceivably improve the epistemic uncertainty~\cite{gal2016uncertainty, jiang2022improving} of the models since they represent rare scenarios. This is particularly important in the case of autonomous driving where replicating these scenarios in the real world might be too dangerous or time consuming.}

\section{Evaluation Setup}
\label{sec:eval-setup}

We evaluate the following \textit{research questions} (RQs):
\begin{itemize}
\item \textbf{RQ1 Effectiveness:} How \textit{effective} is %
\approach %
in 
generating error-inducing inputs that induce class-level fairness violations in
image recognition software?

\item \textbf{RQ2 
Baseline Comparison:} 
How effective is the OOD mutation in comparison to the \textit{baselines}? 

\item \textbf{RQ3 Efficiency:} What is the efficiency (time performance) of \approach 
to generate fairness test cases?

\item \textbf{RQ4 Semantic Validity:} Are the images generated by \approach \textit{semantically valid}, in terms of realism and likelihood of the depicted scenario occurring in real life? Are they comparable to real-world images?

\item \textbf{RQ5 \recheck{Generated images vs. Real-World OOD images}:} \recheck{
What is the model accuracy of our SUT (image classifiers) on 
images generated by \approach 
versus 
real-world OOD images?}

\item \textbf{RQ6 \recheck{Original images vs.  \rev{Error-inducing} OOD images}:}
\revision{What is the accuracy of our SUT (image classifiers) 
on the \rev{error-inducing
images generated by \approach 
versus 
the} corresponding original images from the dataset?
}

\end{itemize}

\noindent
\textbf{Datasets and Subject Programs:} 
We  selected MS-COCO and CityScapes (see \autoref{tab:eval-datasets}) 
due to \revise{the large number of classes (thus, appropriate for testing class-level fairness) present}, 
and their high prevalence in practice (e.g., autonomous cars) and the \reviseCheck{research community~\cite{He_2016_CVPR, xiong2019upsnet, li2021method, treml2016speeding, michaelis2019benchmarking}. 
In particular, we randomly select 300 images containing traffic lights from MS-COCO. For CityScapes, we select 315 road scene images taken in Bremen. Restricting the choice of images in this manner allows us to limit the classes being considered for mutation to the set of objects that are most relevant in road scenes. In the absence of such filtering, we could conceivably insert a car into an image of the sky. Such an insertion would be inappropriate due to the low likelihood of such an situation occurring.}
Additionally, our chosen evaluation subjects (see \autoref{tab:subj-programs}) are the most prominent 
cloud-based image recognition systems %
supporting thousands of objects and scenes.

\noindent
\textbf{Metrics and Measures:} These are defined as follows: %

\begin{itemize}[leftmargin=*]
\item \textbf{Class-level Violations \& Violation Rate:} 
We detect {\em biased} classes via \autoref{eq:fairness-rate}. 
The \textit{class-level violation rate} is the  proportion of biased classes out of all considered classes
(\textit{see} \textbf{RQ1}/\textbf{RQ2}). 

\item \textbf{Error-inducing Inputs \& Fairness Error Rate:} 
We consider a 
generated input (image) to be {\em error-inducing} if 
(1) it leads to an error for a subject and (2) it contributes to the number of 
errors for a class-level violation.
The \textit{fairness error rate} is the proportion of error-inducing inputs 
out of all generated inputs  (\autoref{sec:results}).

\item \textbf{Test Generation Time:} This refers to the time-taken to generate a test suite for class-level group fairness 
(\textit{see} \textbf{RQ3} \autoref{sec:results}).

\end{itemize}

\begin{table}[t]
	\begin{center}
		\caption{Details of Experimental Datasets.}%
	{%
		\resizebox{\linewidth}{!}{
			\begin{tabular}{@{}c|c|c|c|c@{}} %
			\textbf{Dataset} & \textbf{Description} & \textbf{\#Images} & \textbf{\#Classes} & \begin{tabular}{@{}c@{}}\textbf{First} \\ \textbf{Published}\end{tabular}\\
			\hline
			\textbf{MS-COCO}~\cite{lin2014microsoft} & \begin{tabular}{@{}c@{}}Microsoft Common \\ Objects images \end{tabular} & 300 & 183 & 2014 \\ \hline
			\textbf{CityScapes}~\cite{Cordts2016Cityscapes} & \begin{tabular}{@{}c@{}}TU Darmstadt's Urban \\ Street Scenes images \end{tabular} & 315 & 30 & 2015 \\
			\hline
\end{tabular}}}
\label{tab:eval-datasets}   
\end{center}
\end{table}

\begin{table}[t]
\begin{center}
\caption{Details of Subject Programs.}
		\resizebox{\linewidth}{!}{
			\begin{tabular}{@{}l|ccc@{}} %
			\textbf{Subject Programs} & \textbf{\begin{tabular}{@{}c@{}}Description (No. of \\ labels supported)\end{tabular}} & \textbf{\begin{tabular}{@{}c@{}}\#Classes (Our \\ Experiments)\end{tabular}} & \textbf{\begin{tabular}{@{}c@{}}Availability \\ Date\end{tabular}}\\
			\hline
			\textbf{GCP~\cite{google-vision}} & 9000 & 89 & 2017\\ \hline
			\textbf{AWS~\cite{aws-vision}} & 2000+ & 76 & 2016\\ \hline
			\textbf{MS~\cite{ms-vision}} & 10000 & 58 & 2016\\ 
			\hline
	\end{tabular}}
\label{tab:subj-programs}   
\end{center}
\end{table}

\subsection{Research Protocol}
\label{sec:protocol}

We describe the experimental protocol for 
18 different settings (two datasets, three subjects, and three mutations) in our experiments. 

\noindent
\textbf{Clustering and Distribution of Objects:}
For finding the distribution of objects in our initial dataset, we use state-of-the-art library Detectron2~\cite{wu2019detectron2} to detect objects,
whereas 
K-Means algorithm from SciKitLearn~\cite{scikit}
was used for clustering. %
We have selected K-Means since it scales to large data sets, guarantees convergence and 
generalizes to clusters of different shapes and sizes~\cite{kmeans}.

\noindent
\textbf{Image Generation:} %
For all mutations,
\approach attempts to generate one image for each selected class for each given image. %
All experiments except deletion were conducted  five times to 
account for randomness %
in the position, orientation and type of mutated objects. 
Experiments for deletion were 
performed once since deletion is deterministic: %
We delete all objects of a class for all images. 

\noindent
\textbf{Mutated Objects:} 
\revision{
\rev{Due to time constraints, in our experiments, }
\approach applies the mutations to a subset of all the classes present in the dataset. Developers can extend the technique to more classes by obtaining suitable images belonging to each class in question and expending more time to generate OOD images for each class in question. We do, however, consider all the classes present in $\mathbb{C}$ in our fairness error analysis. In particular,} 
\approach attempts to delete or rotate objects belonging to four class labels (namely people, cars, motorcycles, and trucks). 
These classes were selected due to their prevalence in the datasets.
For insertion operation, apart from the four aforementioned classes, we also insert three additional 
classes (namely birds, cats and dogs), as these three classes are common in road scenes.

\noindent
\textbf{Image Caching:} 
We cache the  generated images for test efficiency. %
For a fair  evaluation (\textbf{RQ3}), we only report the results for 
the initial runs for each subject, i.e., MS-COCO using Amazon Rekognition and CityScapes with Google Vision.

\noindent
\textbf{Baseline:} 
\revise{We use two baselines to compare the effectiveness of our OOD mutations: 1) Original data, and 2) ID Mutations. 
In the first case, a developer aims to find class-level fairness violations {\em only} using the original data and not having 
access to \approach.  Meanwhile,} ID mutations aim to transform an image in such a fashion that the distribution of objects 
(i.e., maximum and minimum number of occurrences and the angle of orientations) in the transformed image remains within 
the learned distribution in the respective cluster (i.e., $\mathit{Dist}_{List}$ in Algorithm~\ref{alg:test-gen}). 

\begin{table}[bt!]
  \begin{center}
  \caption{Details of Images in the User Study Dataset.}%
  \resizebox{\linewidth}{!} {%
  \begin{tabular}{@{}l|r|r|rrr@{}} 
  	& \multicolumn{5}{c}{\textbf{\# Images (\# Error-inducing images)}} \\
	\textbf{Dataset} & \textbf{Real}  & \textbf{Mutated} & \textbf{Insertion} & \textbf{Deletion} & \textbf{Rotation}  \\
	\hline
	\textbf{MS-COCO}	& 20 &	20 (17)  & 	6 (6)  & 9 (7)  & 	5 (4)  \\
	\textbf{CityScapes}	& 10  &	10 (7)  &	4 (4)  &	1 (0)  &	5  (3)  \\
	\hline
	\textbf{Total} &	30 	 & 30 (24)  & 	10 (10)  & 	10 (7)  & 	10 (7)  \\	
	 \hline
    \end{tabular}}
  \label{tab:user-study-dataset}   
\end{center}
\end{table}

\noindent
\textbf{Baseline Comparison:}
\revise{For the baseline only using original data, we use the ground truth information (Equation~\ref{eq:gt-reference}) 
to compute the accuracy of each class. Then, the unfair/biased classes are detected as the set of classes whose accuracy 
is below the mean accuracy across all classes (in line with Equation~\ref{eq:fairness-rate})}. 	
For ID mutations, we compare it with the OOD mutation in \approach  for insertion operations. 
This is because most mutated classes in our experiments have \revise{a} minimum object count of zero, thus deleting all 
objects of a class may often generate an image that is ID. Thus, there is no clear boundary between our OOD and an ID 
style deletion operation. Additionally, classes in our dataset have such orientation that any rotation of an object will result 
in an OOD image. Thus, for rotation, the only ID equivalent image is the original image. 
In the insertion experiment, we generate ID images by replacing the OOD constraints in \approach with ID constraints, such that object insertions are 
only performed within the range of the distribution of the object in each cluster. 
However, due to the small range of ID vs. OOD, the generated 
ID inputs beyond first iterations are significantly smaller or already seen in the first. Hence, for a balanced evaluation, we compare only the first iteration 
of \approach with OOD 
to the single run of ID.  
For both OOD and ID, we use the same set of images in the initial datasets with an unlimited time budget. %

\noindent
\textbf{OOD Image Accuracy:}
\reviseThree{We investigate the impact of our OOD style mutations on model accuracy. To this end, we compare the accuracy of 
each SUT on our generated OOD images versus their model accuracy on corresponding real OOD images. In particular, for a given class $c$, we design an experiment to first compute the accuracy on an OOD image $Img$ generated for class $c$. 
This accuracy is then compared with the accuracy on a real image $Img'$ that contains the same number of objects for all classes being considered as in the OOD image $Img$.  \recheck{Similar to the baseline comparisons,  
we also use the insertion mutation operation in this experiment.  
{\em RQ5} presents the results of this experiment. }}

\reviseThree{
\recheck{Specifically,  we implement this experiment as follows: Using the dataset ground truth, ($\mathit{GT}_{Data}(Img)$ from Equation~\ref{eq:gt-reference}),  we determine the number 
	of objects for each considered class present in an original image,  i.e.,  
the classes 
considered for the insertion operation. } 
	We then use the dataset ground truth as a common oracle across all subject programs and for the original images. Additionally,  the oracle is modified accordingly 
	for generated OOD images, specifically, taking into account the number of inserted objects. 
	Algorithm~\ref{alg:ModOOD-oracle} details the procedures involved in implementing this oracle. 
	Given the OOD image, $\mathit{Img}$, 
	the list of classes to be considered, $\mathit{LBL}_{List}$,  the class of object to be inserted, $\mathit{Class}_{Name}$, and the number of objects 
	of said class inserted, $\mathit{Insert}_{Count}$, Algorithm~\ref{alg:ModOOD-oracle} returns the oracle for the number of objects present in 
	the corresponding OOD image ($OOD\_Oracle$). 
	We  restrict the set of objects present to the list of classes being considered (Line~\ref{ln:obj-restrict-start}-Line~\ref{ln:obj-restrict-end}) for insertion. 
	We then find the expected number of objects in the OOD image for each such class 
	(Line~\ref{ln:obj-Inc-start}-Line~\ref{ln:obj-Inc-end}).}
	
	\reviseThree{For each OOD image with respect to a class $c$ ($\in \mathit{LBL}_{List}$), we  identify original images from our dataset that have identical 
	number of objects for all classes in $\mathit{LBL}_{List}$ (according to ground truth $\mathit{GT}_{Data}$). We note that it is possible for multiple images in the dataset to satisfy such 
	a criteria.  Thus, we  
	ensure that each generated OOD image  is paired with exactly one such image (randomly taken) from the dataset. We then compare the accuracy with which the classes being considered, $\mathit{LBL}_{List}$, are detected in the OOD image (using the oracle $OOD\_Oracle$) and the accuracy with which the classes being considered, $\mathit{LBL}_{List}$, are detected in the corresponding original image (using the $\mathit{GT}_{Data}$ oracle). We also note that the total number of objects present in both  $OOD\_Oracle$ and $\mathit{GT}_{Data}$ oracle is necessarily equivalent since we specifically find images that contain the same number of objects.
	} 

\begin{algorithm}[t]
{\scriptsize
		\begin{algorithmic}[1] 
			\Procedure{OOD\_Oracle}{$\mathit{Img},  \mathit{LBL}_{List}, \mathit{Class}_{Name}, \mathit{Insert}_{Count}$}
			
			\LineComment $\mathcal{F}$ returns the dataset ground truth for the image, $\mathit{Img}$  
			
			\State $\mathit{GT}_{Data} \gets \mathcal{F} (\mathit{Img})$

			\State $OOD\_Oracle \gets \emptyset$
			
			\For{$(Obj_L, -) \in \mathit{GT}_{Data}$} \label{ln:obj-restrict-start}
			
				\If {$Obj_L \in \mathit{LBL}_{List}$}
				
				\State $OOD\_Oracle[Obj_L] \gets \mathit{GT}_{Data}[Obj_L]$ \label{ln:obj-restrict-end}
				
				\EndIf
			
			\EndFor
			
			\For{$(Obj_L, -) \in OOD\_Oracle$} \label{ln:obj-Inc-start}
						
				\If {$Obj_L = \mathit{Class}_{Name}$}
			
				\State $OOD\_Oracle[Obj_L] \gets \mathit{GT}_{Data}[Obj_L] + \mathit{Insert}_{Count}$
				
				\Else
			
				\State $OOD\_Oracle[Obj_L] \gets \mathit{GT}_{Data}[Obj_L]$ \label{ln:obj-Inc-end}
				
				\EndIf
				
			\EndFor

			\State \Return $OOD\_{Oracle}$
			
			\EndProcedure
	\end{algorithmic}}
	\caption{OOD oracle for comparison.}
	\label{alg:ModOOD-oracle}
\end{algorithm}

\noindent
\textbf{Fairness Error Analysis:} 
To determine class-level fairness violations,  %
we first filter our classes that are not prominent (occurs <10 times) in our (sub-)datasets to avoid skewness.
 We perform filtering for all experiments, except for the baseline comparison  (\textit{see}  \textbf{RQ2} \autoref{sec:results}). 
 This is due to the relatively smaller set of images involved in baseline (original data and ID generation).

\noindent
\textbf{Implementation Details and Platforms:} 
\approach contains 5K lines of Python code using 
Python 3.7. %
It uses (machine learning and image processing) packages such as PyTorch 1.9, CUDA 110, scikit-learn, numpy and Pillow. 
In addition, we also used the Detectron2~\cite{wu2019detectron2} and LaMa~\cite{suvorov2022resolution} to aid in the image generation. 
For evaluation, we use APIs (with default settings) for each 
of our subject programs (\autoref{tab:subj-programs}). 
All experiments were conducted on a Google Cloud Platform VM %
using an N1 series machine with one vCPU, 20 GB of memory and 
one attached Nvidia Tesla K80 GPU.

\begin{table*}[bt!]
\centering
\caption{Effectiveness of \approach using GT oracle (maximum fairness error rate and violation rate for each (sub)category are in \textbf{bold}). 
}
\resizebox{\linewidth}{!}{%
\begin{tabular}{ccccccccc}
\toprule
\multicolumn{1}{l}{} & \multicolumn{1}{l}{} & \multicolumn{1}{l}{} & \multicolumn{1}{l}{} & {\bf \#Class Violations} & {\bf Violative Rate} & {\bf Error Inputs} & \multicolumn{1}{l}{} & {\bf Fairness Error Rate} \\ \midrule
{\bf Subject} & {\bf Datasets} & {\bf Mutation Ops} & {\bf \#Classes} & {\bf Err Classes} & {\bf Violative Rate} & {\bf Inputs} & {\bf \#Gen Inputs} & {\bf Error Rate} \\
\multirow{6}{*}{GCP} & \multirow{3}{*}{MS-COCO} & Insert & 61 & 16 & 0.26 & 1233 & 6027 & 0.205 \\
 &  & Deletion & 28 & 15 & {\bf 0.54} & 230 & 572 & {\bf 0.402} \\
 &  & Rotation & 49 & 13 & 0.27 & 317 & 1425 & 0.222 \\
\cmidrule(l){3-9}
 & \multirow{3}{*}{CityScapes} & Insert & 22 & 2 & 0.09 & 377 & 8486 & 0.044 \\
 &  & Deletion & 14 & 4 & 0.29 & 99 & 620 & {\bf 0.160} \\
 &  & Rotation & 19 & 8 & {\bf 0.42} & 297 & 2500 & 0.119 \\
\midrule
\multirow{6}{*}{MS} & \multirow{3}{*}{MS-COCO} & Insert & 46 & 16 & 0.35 & 2895 & 5908 & 0.490 \\
 &  & Deletion & 19 & 7 & 0.37 & 253 & 572 & 0.442 \\
 &  & Rotation & 31 & 14 & {\bf 0.45} & 762 & 1425 & {\bf 0.535} \\
\cmidrule(l){3-9}
 & \multirow{3}{*}{CityScapes} & Insert & 17 & 7 & {\bf 0.41} & 3305 & 9085 & {\bf 0.364} \\
 &  & Deletion & 10 & 4 & 0.40 & 182 & 620 & 0.294 \\
 &  & Rotation & 17 & 5 & 0.29 & 493 & 2500 & 0.197 \\
\midrule
\multirow{6}{*}{AWS} & \multirow{3}{*}{MS-COCO} & Insert & 35 & 6 & 0.17 & 716 & 4583 & 0.156 \\
 &  & Deletion & 17 & 8 & {\bf 0.47} & 96 & 572 & {\bf 0.168} \\
 &  & Rotation & 27 & 10 & 0.37 & 124 & 1425 & 0.087 \\
\cmidrule(l){3-9}
 & \multirow{3}{*}{CityScapes} & Insert & 15 & 2 & 0.13 & 139 & 7104 & 0.020 \\
 &  & Deletion & 11 & 5 & {\bf 0.45} & 293 & 620 & {\bf 0.473} \\
 &  & Rotation & 13 & 1 & 0.08 & 18 & 2500 & 0.007 \\
\midrule
\multirow{2}{*}{Dataset} & MS-COCO & All & 313 & 105 & {\bf 0.34} & 6626 & 22509 & {\bf 0.294} \\
 & CityScapes & All & 138 & 38 & 0.28 & 5203 & 34035 & 0.153 \\
\midrule
\multirow{3}{*}{Subject} & GCP & All & 193 & 58 & 0.30 & 2553 & 19630 & 0.130 \\
 & MS & All & 140 & 53 & {\bf 0.38} & 7890 & 20110 & {\bf 0.392} \\
 & AWS & All & 118 & 32 & 0.27 & 1386 & 16804 & 0.082 \\
\midrule
Total & All & All & 451 & 143 & 0.32 & 11829 & 56544 & 0.209 \\ \bottomrule
\end{tabular}%
}
\label{tab:effectivenessGT}
\end{table*}

\subsection{User Study Design}
\label{sec:study-design}
Our study had 105 users and 60 images 
to examine if the generated images are \textit{realistic}  to humans and 
\textit{likely to occur in the real world}.

\noindent
\textbf{Study Dataset:} 
We first randomly selected 30 mutated images from \approach such that all three mutation 
operators were equally represented. We also took additional care to ensure that images from both datasets were included. 
In particular, we selected 20 images from MS-COCO (vs 10 from CityScapes) due its %
significantly larger number of class labels in comparison to CityScapes 
(183 vs. 20, \textit{see} \autoref{tab:eval-datasets}). We also ensure that most of the selected images (24 out of 30) 
induce errors for at least one subject program (\textit{see} \autoref{tab:user-study-dataset}). 
We then select the corresponding 30 real images for comparison.

\noindent
\textbf{Survey Questionnaire:}
We provide participants a randomly ordered set of 60 images in our study dataset. To avoid bias, we ensure that all consecutive 
images do not have the same mutation operation, and a mutated image is not next to its corresponding original image. To validate 
the soundness of participant responses, we ask participants to also provide the number of vehicles in each image. 
Specifically, \revise{the} following questions were posed: 
\begin{itemize}[leftmargin=*]
\item \textbf{Image Realism:} ``On a scale of 0 to 10, how realistic is the image? ``Realistic'' means the image depicts or seems to depict real people, objects or scenarios.''
\item \textbf{Scene Likelihood:} ``On a scale of 0 to 10, how likely is the scenario depicted in the image to occur in real life?''
\item \textbf{Validation:} ``How many vehicles (e.g., cars) are in this image?''
\end{itemize}

The questionnaire is available here: 
\url{https://bit.ly/3B1Qc12}

\noindent
\textbf{Participants:} We conducted this study on Amazon Mechanical Turk (MTurk)~\cite{mturk}. We received 105 responses  
in 11.25 hours. Each participant took about 66 minutes to complete the study, on average. 

\noindent
\textbf{Response Data Validation:} We validated 81 responses by checking the answers for  the number of vehicles in the images. 
We randomly chose five unambiguous images (with few and clear number of vehicles) for validation. 
We also ensured that the user agreement 
for the number of vehicles in the images is high (above 75\%). Then we set a 60\% (3 out 5) correctness threshold for these 
five images. %

\noindent
\textbf{Response Data Analysis:} To determine semantic validity of our images, we collated the Likert scale scores for the 81 
valid responses using the two questions on the realism of the images and the likelihood of the depicted scenarios. We analyse 
semantic validity using both scores for original versus mutated images across different mutations, datasets and 
error-inducing images (see \textbf{RQ4} in \autoref{sec:results}).

\section{Evaluation Results}
\label{sec:results}

\noindent
\textbf{RQ1 Effectiveness:}
\revise{We evaluate the effectiveness of \approach using both the GT and MT oracle (\autoref{tab:effectivenessGT} and \autoref{tab:effectiveness}),  
as discussed in \autoref{sec:overview}. The choice of GT test oracle is useful when practitioners have access to ground 
truth information on the dataset, whereas the MT oracle is useful when practitioners %
lack access to detailed information on the dataset or other subject programs.}

\noindent
\textbf{\textit{Using Ground Truth Oracle:}} 
\revise{
\autoref{tab:effectivenessGT} shows that 
\textit{\approach is effective in exposing class-level fairness violations using ground truth information.} In particular, \approach reveals 32\% class-level fairness violations w.r.t. ground truth information. About \textit{one in five inputs} (21\%) generated by \approach exposed a class-level fairness violation. 
We observed that MS-COCO dataset and the Microsoft Vision subject program are more error-prone than other datasets (i.e., CityScapes) and other subject programs (GCP and AWS). For instance, \approach exposed more fairness violations (0.34 vs. 0.28) and generated more error-inducing inputs (0.294 vs. 0.153) for MS-COCO than CityScapes (\textit{see} \autoref{tab:effectivenessGT}). Overall, \approach effectively exposes 
class-level fairness violations with GT oracle. 
}

\begin{result}
\revise{
21\% of the OOD images generated by 
\approach 
reveal class-level fairness violations in 
32\% of classes, using ground truth oracle. 
}
\end{result}

\begin{table*}[t]
\centering
\caption{Effectiveness of \approach  using MT oracle (maximum fairness error rate and violation rate for each (sub)category are in \textbf{bold}). {\bf Ex.:} Exclusion, {\bf Inc.:} Inclusion. 
}
\resizebox{\linewidth}{!}{
\begin{tabular}{@{}ccccccccccccc@{}}
\toprule
 &  &  & & \multicolumn{2}{c}{\textbf{\#ClassViolations}} & \multicolumn{2}{c}{\textbf{Violative Rate}} & \multicolumn{2}{c}{\textbf{\#Error-inducing inputs}} &  & \multicolumn{2}{c}{\textbf{Fairness Error Rate}} \\ \midrule
\textbf{Subject} & \textbf{Datasets} & \textbf{Mutation Ops} & \textbf{\#Class} & \textbf{Ex.} & \textbf{Inc.} & \textbf{Ex.} & \textbf{Inc.} & \textbf{Ex.} & \textbf{Inc.} & \textbf{\#gen-Inputs} & \textbf{Ex.} & \textbf{Inc.} \\
 &  & Insertion & 61 & 13 & 12 & 0.21 & 0.2 & 354 & 1072 & 6027 & 0.059 & 0.178 \\
 &  & Deletion & 23 & 9 & 7 & {\bf 0.39} & {\bf 0.3} & 124 & 193 & 572 & {\bf 0.217} & {\bf 0.337} \\
 & \multirow{-3}{*}{MS-COCO} & Rotation & 49 & 12 & 8 & 0.24 & 0.16 & 144 & 240 & 1425 & 0.101 & 0.168 \\ \cmidrule(l){3-13} 
 &  & Insertion & 22 & 11 & 2 & 0.5 & 0.09 & 2354 & 418 & 8486 & {\bf 0.277} & 0.049 \\
 &  & Deletion & 11 & 4 & 4 & 0.36 & {\bf 0.36} & 159 & 195 & 620 & 0.256 & {\bf 0.315} \\
\multirow{-6}{*}{GCP} & \multirow{-3}{*}{CityScapes} & Rotation & 18 & 7 & 4 & {\bf 0.39} & 0.22 & 269 & 50 & 2500 & 0.108 & 0.020 \\ \midrule
 &  & Insertion & 43 & 16 & 5 & 0.37 & 0.12 & 1731 & 1640 & 5908 & 0.293 & 0.278 \\
 &  & Deletion & 14 & 6 & 6 & {\bf 0.43} & {\bf 0.43} & 221 & 192 & 572 & {\bf 0.386} & 0.336 \\
 & \multirow{-3}{*}{MS-COCO} & Rotation & 29 & 10 & 8 & 0.34 & 0.28 & 348 & 672 & 1425 & 0.244 & {\bf 0.472} \\ \cmidrule(l){3-13}
 &  & Insertion & 17 & 7 & 5 & {\bf 0.41} & 0.29 & 416 & 2712 & 9085 & 0.046 & {\bf 0.299} \\
 &  & Deletion & 9 & 3 & 3 & 0.33 & {\bf 0.33} & 129 & 70 & 620 & {\bf 0.208} & 0.113 \\
\multirow{-6}{*}{MS} & \multirow{-3}{*}{CityScapes} & Rotation & 17 & 3 & 5 & 0.18 & 0.29 & 213 & 291 & 2500 & 0.085 & 0.116 \\ \midrule
 &  & Insertion & 34 & 9 & 1 & 0.26 & 0.03 & 245 & 156 & 4583 & 0.053 & 0.034 \\
 &  & Deletion & 16 & 5 & 6 & {\bf 0.31} & {\bf 0.38} & 74 & 49 & 572 & {\bf 0.129} & 0.086 \\
 & \multirow{-3}{*}{MS-COCO} & Rotation & 26 & 7 & 6 & 0.27 & 0.23 & 100 & 223 & 1425 & 0.070 & {\bf 0.156} \\ \cmidrule(l){3-13}
 &  & Insertion & 15 & 7 & 2 & {\bf 0.47} & 0.13 & 842 & 3886 & 7104 & 0.119 & {\bf 0.547} \\
 &  & Deletion & 11 & 4 & 4 & 0.36 & {\bf 0.36} & 300 & 27 & 620 & {\bf 0.484} & 0.044 \\
\multirow{-6}{*}{AWS} & \multirow{-3}{*}{CityScapes} & Rotation & 13 & 3 & 1 & 0.23 & 0.08 & 566 & 18 & 2500 & 0.226 & 0.007 \\ \midrule
 & MS-COCO & All & 295 & 87 & 59 & 0.29 & 0.20 & 3341 & 4437 & 22509 & 0.148 & 0.197 \\
\multirow{-2}{*}{Dataset} & CityScapes & All & 133 & 49 & 30 & {\bf 0.37} & {\bf 0.23} & 5248 & 7667 & 34035 & {\bf 0.154} & {\bf 0.225} \\ \midrule
 & GCP & All & 184 & 56 & 37 & 0.30 & 0.20 & 3404 & 2168 & 19630 & {\bf 0.173} & 0.110 \\
 & MS & All & 129 & 45 & 32 & {\bf 0.35} & {\bf 0.25} & 3058 & 5577 & 20110 & 0.152 & {\bf 0.277} \\
\multirow{-3}{*}{Subject} & AWS & All & 115 & 35 & 20 & 0.30 & 0.17 & 2127 & 4359 & 16804 & 0.127 & 0.259 \\ \midrule
Total & All & All & 428 & 136 & 89 & 0.32 & 0.21 & 8589 & 12104 & 56544 & 0.152 & 0.214 \\ \bottomrule
\end{tabular}}
\label{tab:effectiveness}  
\end{table*}

\noindent
\textbf{\textit{Using Metamorphic Oracle:}}
\revise{
\autoref{tab:effectiveness} shows that 
\approach revealed 32\% class-level fairness violations relating to exclusion errors and 21\% class-level fairness violations for inclusion 
errors. %
In addition, we observed that up to one in five inputs generated by our approach reveals a class-level fairness violation. 
For instance, 21\% of the generated inputs %
exposed class-level fairness violations relating to inclusion errors 
across all settings 
(\textit{see} \autoref{tab:effectiveness}). Although \approach is effective across all settings,
we found that it finds more errors using CityScapes dataset %
than using MS-COCO. 
We attribute the effectiveness %
to the use of distribution-aware mutations, which drive the input generation to induce class-level fairness violations.  
}

\begin{result}
\revise{
Using the metamorphic oracle, 
one-fifth of the inputs generated by \approach %
revealed fairness errors 
in one-third 
of classes. %
}
\end{result}

\begin{figure}[t]
	\centering
	\begin{subfigure}[t]{.25\textwidth}
		\centering
		\includegraphics[scale=0.32]{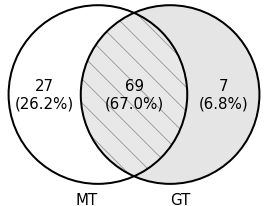}
		\caption{Unfair Classes found by \\ GT vs. MT oracles.}
	\label{fig:venn-MT-GT}
\end{subfigure}%
\begin{subfigure}[t]{.22\textwidth}
	\centering
	\includegraphics[scale=0.32]{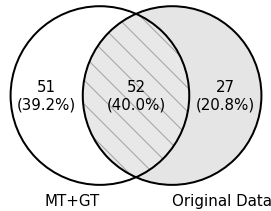}
	\caption{Unfair classes (GT \& MT Oracles vs. Original Data).}
	\label{fig:venn-Org-MTGT}
\end{subfigure}
\caption{%
	Illustration of \approach effectiveness.}
\label{fig:venn-diagrams}
\end{figure}

\noindent
\textit{\textbf{Test Oracle Comparisons:}}
\revise{
\autoref{fig:venn-MT-GT}  illustrates that the MT oracle 
exposed most (91\% =69/76) of the fairness violations found by the GT oracle. 
Besides, \textit{two-third (67\% = 69/103) of all violated classes are found by both oracles}. %
We also observed that \textit{almost 7\% the violated classes found by the GT oracle are missed by the MT oracle}. 
This is due to the difference in the number of classes identified by both oracles. In our setting, the GT oracle identifies 
more classes than the MT oracle, since it obtains image recognition data from multiple sources (i.e., all subjects and 
dataset labels) in comparison to the MT
oracle (a single subject). This 
directly 
influences the mean error rate and the found violated classes. 
Finally, we observed that the \textit{MT oracle exposed 26\% of fairness violations that are missed by the GT oracle}. 
Unlike the GT oracle, the MT oracle accounts for %
errors where the subject performs better on the mutated image (e.g., AWS in \autoref{tab:approach-overview}). 
This is useful to expose weaknesses in a subject. 
}
\begin{result}
\revise{
The MT oracle is a good (proxy) estimator 
of the GT oracle. MT revealed most (69/76 $\approx$ 91\%) of the fairness violations found by GT. 
}
\end{result}

\begin{table*}[bt!]
	\centering
	\caption{ Comparison of \approach , i.e., \textit{out-of-distibution} (OOD) mutation-based fairness test generation approach to the baseline, i.e., \textit{in-distribution} (ID) mutation-based fairness test generation. {\bf Ex.:} Exclusion, {\bf Inc.:} Inclusion.
	}
	\resizebox{\linewidth}{!}{
		\begin{tabular}{@{}ccccccccccccc@{}}
			\toprule
			&  &  &  & \multicolumn{2}{c}{\textbf{\#ClassViolations}} & \multicolumn{2}{c}{\textbf{Violative Rate}} & \multicolumn{2}{c}{\textbf{\#Error-inducing inputs}} &  & \multicolumn{2}{c}{\textbf{Fairness Error Rate}} \\ \midrule
			\textbf{Distribution} & \textbf{Subject} & \textbf{Datasets} & \textbf{\#Class} & \textbf{Ex.} & \textbf{Inc.} & \textbf{Ex.} & \textbf{Inc.} & \textbf{Ex.} & \textbf{Inc.} & \textbf{\#gen-Inputs} & \textbf{Ex.} & \textbf{Inc.} \\
			& GCP & All & 81 & 13 & 14 & 0.16 & 0.17 & 24 & 121 & 1100 & 0.022 & 0.11 \\
			& MS & All & 58 & 17 & 11 & 0.29 & 0.19 & 144 & 222 & 917 & 0.157 & 0.242 \\
			\multirow{-3}{*}{ID} & AWS & All & 54 & 7 & 4 & 0.13 & 0.07 & 14 & 161 & 1355 & 0.01 & 0.119 \\ \midrule
			& GCP & All & 78 & 25 & 11 & 0.32 & 0.14 & 503 & 466 & 2896 & 0.174 & 0.161 \\
			& MS & All & 56 & 23 & 11 & 0.41 & 0.2 & 453 & 834 & 3005 & 0.151 & 0.278 \\
			\multirow{-3}{*}{\approach} & AWS & All & 37 & 8 & 4 & 0.22 & 0.11 & 73 & 92 & 2343 & 0.031 & 0.039 \\ \midrule
			ID & All & All & 193 & 37 & 29 & 0.19 & 0.15 & 182 & 504 & 3372 & 0.054 & 0.149 \\
			\approach & All & All & 171 & 56 & 26 & 0.33 & 0.15 & 1029 & 1392 & 8244 & 0.125 & 0.169 \\ \midrule
			\multicolumn{3}{c}{Improvement (\%)} & NA & NA & NA & 73.68 & 0 & NA & NA & NA & 131.48 & 13.42 \\ \bottomrule
	\end{tabular}}
	\label{tab:baseline-comparison}
\end{table*}

\noindent
\textbf{RQ2 Baseline Comparison:}
\revise{
We compare \approach to fairness analysis with (a) 
only \textit{original data} (\approach vs. Original Data) and (b) only \textit{in-distribution} (ID) mutation (\approach vs. ID).
}

\noindent
\textbf{\textit{\approach vs. Original Data:}}
\revise{In this experiment, we consider an approach with developers inspecting fairness violations {\em only} in the 
original dataset and without access to  OOD test suite. \autoref{fig:venn-Org-MTGT} highlights the similarity and 
differences in the class-level fairness violations exposed by such an approach with respect to \approach.}

\revise{
We found that \textit{\approach exposes (30\%) more class-level fairness violations than the original data (103 vs. 79)} (\textit{see} \autoref{fig:venn-Org-MTGT}). More importantly, a developer using only the original dataset will miss 39.2\% (51 out of 130) of the class-level fairness violations exposed.
In addition, \approach is a good proxy for determining the class-level fairness violations found in the original dataset, since it exposes 66\% (52 out of 79) of the class-level fairness violations exposed by the original dataset. 
These results highlight the need for generating OOD data, as they demonstrate that 
\approach is effective in exposing fairness violations %
missed by the original dataset. 
}

\reviseCheck{
In addition, we performed a statistical analysis on both sets of images. We calculate the error rates for each image based on our ground truth reference, $\mathit{GT}_{Ref}$, taking care to exclude the class being mutated from the error calculation. More specifically, we conducted a Mann–Whitney U-test to determine whether the original images from the dataset were distinguishable from generated OOD images using their error rates.
The Mann-Whitney U test shows that there is a significant difference between the two sets of images; it yields a test statistic of 41951407.5 and a p-value of $\approx$ $1 \cdot 10^{-46}$
}

\begin{result}
\revise{
Class-level fairness analysis with \approach is more effective than using only the original dataset. 
\approach exposes (30\%) more class-level fairness violations than the original dataset and 39.2\% of all found violations were exposed by \approach only. 
}
\end{result}

\noindent
\textbf{\textit{\approach vs. ID:}}
In this experiment, we compare the 
OOD style mutation of \approach with the alternative \textit{in-distribution} (ID) 
mutation. 
As discussed in \autoref{sec:protocol} (Baseline Comparison), we employ only the insertion operation for this 
comparison. 
 
Our evaluation results show that \textit{OOD style mutation outperforms the ID-based mutation approach in revealing class-level group 
fairness violations} (see \autoref{tab:baseline-comparison}). 
Specifically, \approach reveals up to 74\% more class-level fairness violations than the baseline for exclusion errors (\textit{see} \autoref{tab:baseline-comparison}). 
In addition, we found that a developer is more than two times likely (up to 131\%) to find class-level fairness errors with OOD than ID. 
Furthermore, OOD generates over 8K inputs and 1029 error-inducing inputs for exclusion errors, while ID generates 
only 182 error-inducing inputs and over 3K total inputs.  This is particularly due to the fact that the input space for 
OOD is typically much larger than ID, since ID mutations are constrained within a static range.  These results suggest 
that our use of OOD-based mutation contributes significantly to the effectiveness of \approach.

\reviseCheck{
Similarly, we conducted a Mann–Whitney U-test to 
determine whether the generated in-distribution images were distinguishable from generated OOD images when using their error rates. We found that there is a significant difference between the two sets of images; it yields a test statistic of 86298859.0 and a p-value of $\approx$ $2 \cdot 10^{-20}$.
}

\begin{result}
	OOD %
	mutation significantly 
	contributes to \approach's effectiveness. It is up to 2.3X as effective 
	as ID mutation. 
\end{result}

\noindent
\textbf{RQ3 Efficiency:}
\reviseThree{This RQ examines the efficiency (time performance) of our approach (\approach) in 
generating fairness test suites.  
For a fair and balanced evaluation, we only report the time-taken for \approach during initial execution without 
\textit{caching the generated images} for each dataset.
Hence, we report the time-taken for the two initial experimental settings with MSCOCO using Amazon Rekognition
 (aka AWS) and the CityScapes dataset with Google Vision API (aka GCP).}
 
\autoref{tab:Efficiency-Table} reports the  test generation time of \approach. 
It highlights that the two initial experimental setups took about 39 hours to complete the 
generation of 18K inputs. %
This implies that \approach generates a fairness test case in about 7.7 seconds, on average. 
Moreover, the number of exposed fairness violations and generated error-inducing inputs within the test generation time 
is reasonable for a developer. For instance, 
\approach  generated hundreds (847) of error-inducing inputs and exposed 34 class-level fairness violations 
within 15 hours of fairness test generation, when testing AWS using the MS-COCO dataset (\textit{see} \autoref{tab:effectiveness}). 
Further inspection shows that 
these results hold across mutation operations. %
In particular, the deletion operation is the fastest mutation operation (about 6.7 seconds)  and the rotation operation is the most 
expensive operation (10 seconds), on average. 
Deletion operation is cheaper due to the single deterministic attempt at 
deleting all objects of the class in the image. In contrast, rotation  is more expensive 
since it requires inpainting and insertion. 
The performance of \approach across the datasets is similar. Specifically, \approach took about 
7.5-8 seconds to generate an input across both datasets. 
We attribute this efficiency to the lightweight and inexpensive nature of our distribution-aware mutation operations. 

\begin{result}
\revision{On average, \approach 
takes $\approx$ 7.7 sec 
to generate a test.}
\end{result}

\begin{table}[t]
\caption{ Test Generation Efficiency of \approach. %
}
\resizebox{\linewidth}{!}{
\begin{tabular}{@{}ccccc@{}}
\toprule
\multicolumn{1}{l}{\textbf{}} & \multicolumn{4}{c}{\textbf{Time Taken in seconds (\#Images Generated)}} \\ \midrule
\textbf{Dataset (subject)} & \textbf{Insertion} & \textbf{Deletion} & \textbf{Rotation} & \textbf{Total} \\
\textbf{MS-COCO (AWS)} & 38112 (4583) & 2698 (572) & 12502 (1425) & 53312 (6580) \\
\textbf{CityScapes (GCP)} & 54518 (8486) & 5257 (620) & 27118 (2500) & 86893 (11606) \\ \midrule
\textbf{Total} & 92630 (13069) & 7955 (1192) & 39620 (3925) & 140205 (18186) \\ \bottomrule
\end{tabular}}
\label{tab:Efficiency-Table}
\end{table}

\begin{table*}[bt!]
  \begin{center}
  \caption{Semantic validity (realism and likelihood) of real images versus \approach's generated images.
  } %
  {\footnotesize
  \resizebox{\linewidth}{!}{
  \begin{tabular}{@{}|l|r|r|rrr|r|r|rrr|@{}} %
  \hline
  	& \multicolumn{10}{c|}{\textbf{Semantic Validity of All Images (only Error-inducing images)}} \\
  	& \multicolumn{5}{c|}{\textbf{Realism of Images}} & \multicolumn{5}{c|}{\textbf{Likelihood of Scenarios}} \\
	\textbf{Dataset} & \textbf{Real}  & \textbf{Mutated} & \textbf{Insertion} & \textbf{Deletion} & \textbf{Rotation}  &  \textbf{Real} & \textbf{Mutated} & \textbf{Insertion} & \textbf{Deletion} & \textbf{Rotation}   \\
	\hline
	\texttt{MS-COCO}	& 7.83 & 6.56 (6.52)  &  5.66 (5.66)  & 7.14 (6.99)  & 6.59 (6.99)  & 8.08  &  6.89 (6.85)  & 6.12 (6.12)  &  7.44 (7.31) & 6.81 (7.16)  \\
		\texttt{CityScapes}	& 8.02 & 5.93 (5.53)  &  4.67 (4.67)  & 7.84 (NA)  & 6.56 (6.67)  &8.12  &  6.36 (6.03)  & 5.28 (5.28)  &  7.79 (NA) & 6.95 (7.04) \\
	\hline
	\textbf{Total} & 7.89  & 6.35 (6.23)  & 5.26 (5.26)  & 7.21 (6.99)  & 6.57 (6.85)  & 8.11   &  6.71 (6.61)  & 5.78 (5.78)  &  7.48 (7.31) & 6.88 (7.11) \\	
	\hline
		\textbf{Real vs. Mut (\%)} & NA & 80.4	(78.9) & 66.7	(66.7) & 91.4	(88.6) & 83.3	(86.7) & NA & 82.8 (81.6) & 71.3	(71.3) & 92.2	(90.1) & 84.9	(87.7)			\\
		\hline
    \end{tabular}}}
  \label{tab:user-study-results}   
\end{center}
\end{table*}

\noindent
\textbf{RQ4 Semantic Validity:}
\rev{To measure validity, we conduct two experiments, namely (1) a qualitative user study, and (2) a quantitative image quality experiment. Both of which are accompanied with statistical analysis. 
In the following, we report the settings and results of each experiment. }

\noindent
\textbf{\rev{User Study:}}
\rev{Firstly,  } we conducted a \textit{user study} to evaluate the \textit{semantic validity} of the images generated by \approach. Our study involves 105 participants and 60 images (see \autoref{sec:study-design}). 
\revision{\rev{
This qualitative user study 
allows to accurately capture the realism of our images, especially from the human perspective. } We note that a user study continues to be common practice in assessing the effectiveness of the generated images~\cite{gadde2021detail, lin2018st, fan2014automated}. In addition, most methods used to assess image quality are based on learning-based models where a quality prediction model is learned from data \rev{that is labeled by humans}~\cite{ma2017end, jin2022pseudo}.}

Our user study results show that \textit{images generated by our test generator (\approach) are semantically valid, %
when compared to real-world images}.
\autoref{tab:user-study-results} shows that our mutation operations are (up to 91\%) as realistic 
as real-world images and (up to 92\%) likely to occur in real life (\textit{see ``Real vs. Mut'' deletion operation}). 
We observed that the deletion operation produces the most (up to 92\%) semantically valid images. Meanwhile, the insertion operation produces 
the least realistic images, yet images resulting from the insertion operation are (up to 71\%) likely to occur in real life. We also observed that these 
results are similar for the error-inducing images, i.e., images that cause an error in at least one subject program.  
Furthermore, we found that both benign and error-inducing images were seen as being similarly valid, realistic and likely to occur.
Overall results show that all tested images generated by \approach are 80\% as realistic as real-world images. Participants 
also report that generated images depict scenarios that are 
83\% as likely to occur in real life when compared to the original images. 
This suggests that the OOD images generated 
by \approach do not deviate significantly from real-world 
expectations of humans. 
Additionally, such results hold regardless of 
the error-inducing ability of the images and type of mutation operators. 

\begin{figure}
	\centering
	\includegraphics[scale=0.6]{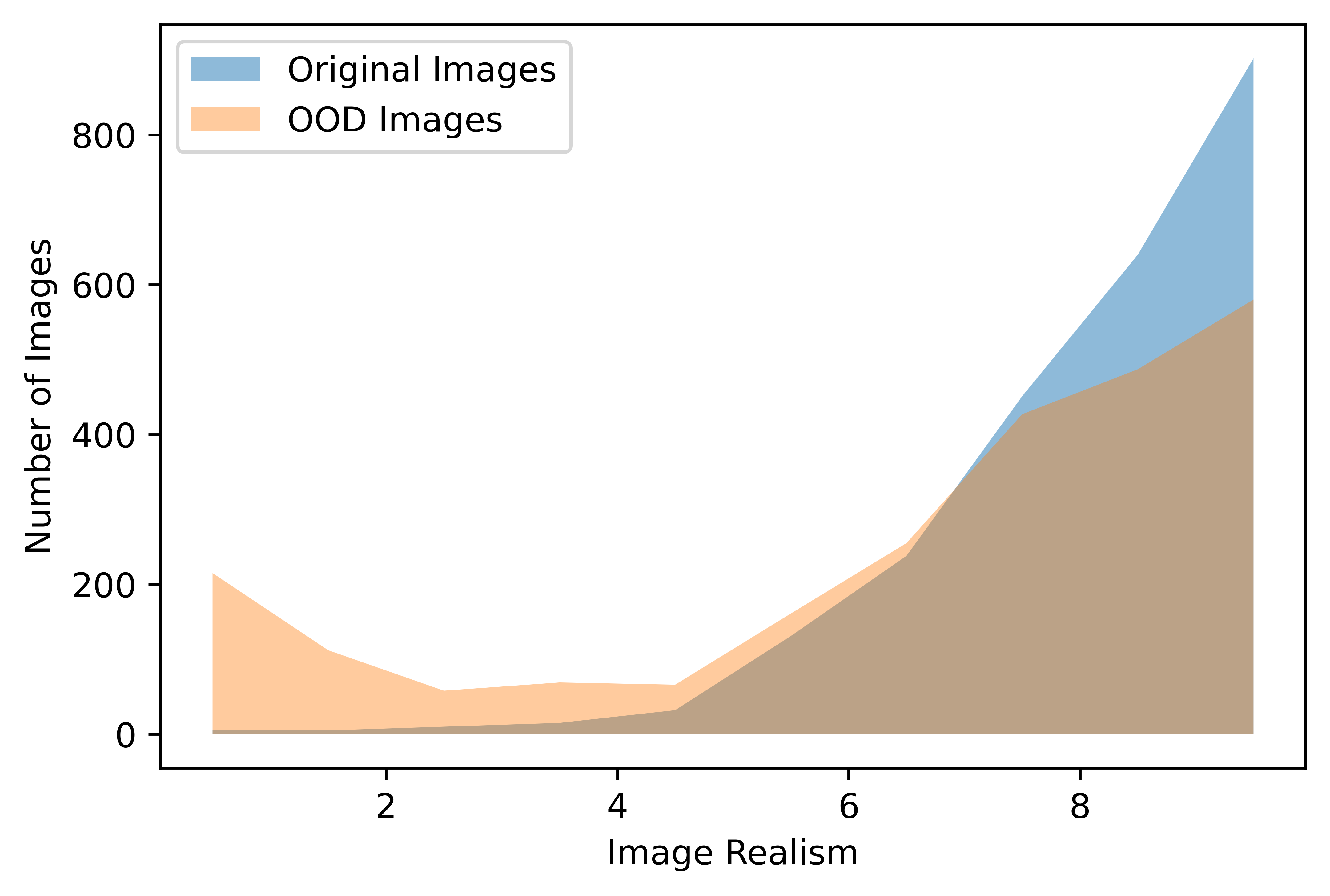}
	\caption{Distribution of the number of images with a given realism score, where 1 is the lowest possible score and 10 is the highest possible score.
	}
	\label{fig:realOverlap}
\end{figure}

\noindent
\textbf{\rev{Statistical Analysis (User Study):}}
\reviseThree{
\recheck{In addition,  we performed a statistical analysis 
of our user study results.  Specifically,  we conducted a Mann–Whitney U-test to determine 
	whether the original images from the dataset were indistinguishable from their corresponding OOD images using the reported realism scores of participants. 
The Mann-Whitney U test 
shows that there is a clear difference in the realism of the two sets of images,  it 
yields 
	a test statistic of 3791661.5 and a p-value of zero.  \autoref{fig:realOverlap} also provides an 
	overlapping frequency graph of the two sets of scores.  It shows that the scores for both real and OOD images 
	mostly overlap.  However, we also observe that a portion of our OOD images have scores that are very low (between one and three inclusive) 
	indicating that some of our generated images might be unrealistic.  This is primarily due to the current limitations of the state-of-the-art software tools for image mutations (e.g.,  current object insertions techniques are unable to perfectly blend inserted objects into the original image). For instance, inserted objects might not perfectly match the lighting conditions present in the original image. As such tools become more mature in the future,  we expect to obtain more %
	realistic images using \approach and our mutation operations.}
}

\noindent
\textbf{\rev{Image Quality Analysis:}}
\revision{
\rev{Finally, we }
quantitatively evaluate the overall quality of the generated images \rev{using} 
the PyTorch Image Quality (PIQ) \rev{library}~\cite{kastryulin2022piq}. In particular, we evaluate the image quality of the set of original images versus OOD images (generated by \approach) in our user study using CLIP-IQA~\cite{wang2023exploring}. \rev{CLIP-IQA evaluates the quality of an image 
using two antonym prompts}.  
\rev{Antonym prompts allows it to
to accurately determine} where on the spectrum a particular image falls. In our experiments, we use CLIP-IQA to evaluate the 
quality of the image by providing it with the custom prompts, 
\rev{``Realistic photo'' and ``Unrealistic photo''}. 
This allows us to evaluate the realism of the image. 
}

\revision{
\rev{Our evaluation results show that 
the quality of the set of OOD images, generated by \approach, are similar 
to that of the original images. 
The CLIP-IQA score for the original images is
0.695,
while the CLIP-IQA score for the OOD images, generated by \approach, is
0.696,
on average.  Using the 
Mann-Whitney U test,  we also perform statistical analysis to determine whether the two sets of images can be differentiated on the basis of the CLIP-IQA score.}  \rev{We found that the 
quality
 of the set of OOD images, generated by \approach, are statistically indistinguishable from the original images. In particular,  the Mann-Whitney U test yields a test statistic of 
448.5
and a p-value of 
0.99.
These results show that \approach and its mutation preserves the image quality in the original images}. 
}

\begin{result}
Generated images %
are (up to 91\%) as realistic as real images, and 
the depicted scenes are (up to 92\%) likely to occur in real life.

\end{result}

\noindent
\textbf{RQ5 \recheck{Generated images vs. Real-World OOD image}:}
\reviseThree{In this experiment, we compare the accuracy of our SUTs on 1) OOD images generated by the insertion operation in \approach, versus 2) images from the dataset that contain equivalent number of objects present.  \autoref{tab:genOODvsRealOOD} highlights our results. }

\reviseThree{
We found that 
the SUT accuracy on OOD images generated by \approach is better than the accuracy on real images containing similar number of objects (\textit{see} \autoref{tab:genOODvsRealOOD}).  
\recheck{In particular,  we note that AWS achieves similar accuracy for both generated images and real-world OOD images from the CityScapes dataset: In this case, the model accuracy on the images generated by \approach are only 
5.4\% more than that of real-world 
OOD images. } In addition, the OOD images produced by \approach had a consistently higher accuracy when compared to the real images from the dataset. The difference in the accuracy can be attributed to our inserted objects being placed 
in focus, e.g.,  
because our insertion operation does not 
place objects behind already existing objects.  \recheck{Notably,  images with object occlusion,  which may be present in the images in the dataset,  are more difficult to detect.  However,  for our approach,  the increased likelihood of generating images %
without occlusion leads to much higher detection rates.  
Overall,  these results indicate that (a) our image mutations do not adversely impact the accuracy of image recognition software,  and (b) the errors exposed by \approach are not due to performance degradation in the recognition of the image classifiers.  
This is evident for 
GCP where SUT accuracy on generated images is 
2.7x as much as on 
real OOD images in the dataset. }}

\begin{result}
\recheck{
SUT accuracy on OOD images generated by \approach  
is (up to 170\%) more than  
real-world OOD images.}
\end{result}

\begin{table}[]
\centering
\caption{Accuracy of our SUTs on OOD images generated by \approach and similar images from the dataset.}
\resizebox{\linewidth}{!}{%
\begin{tabular}{@{}ccccccc@{}}
\toprule
\multicolumn{1}{l}{} & \multicolumn{3}{c}{\textbf{MSCOCO}}       & \multicolumn{3}{c}{\textbf{CityScapes}}   \\
\multicolumn{1}{l}{} &
  \multicolumn{1}{c}{\textbf{\begin{tabular}{@{}c@{}}Dataset \\ Images \end{tabular}}} &
  \multicolumn{1}{c}{\textbf{\begin{tabular}{@{}c@{}}OOD \\ Images \end{tabular}}} &
  \multicolumn{1}{c}{\textbf{\begin{tabular}{@{}c@{}}OOD Acc. \\ Imp. (\%) \end{tabular}}} &
  \multicolumn{1}{c}{\textbf{\begin{tabular}{@{}c@{}}Dataset \\ Images \end{tabular}}} &
  \multicolumn{1}{c}{\textbf{\begin{tabular}{@{}c@{}}OOD \\ Images \end{tabular}}} &
  \multicolumn{1}{c}{\textbf{\begin{tabular}{@{}c@{}}OOD Acc. \\ Imp. (\%) \end{tabular}}} \\
\midrule
\textbf{GCP}         & 0.180 & 0.290 & 60.9 & 0.132 & 0.358 & 170.6 \\
\midrule
\textbf{MS}          & 0.182 & 0.284 & 56.2 & 0.138 & 0.275 & 99.3 \\
\midrule
\textbf{AWS}         & 0.584 & 0.736 & 26.0 & 0.602 & 0.635 & 5.4 \\
\bottomrule
\end{tabular}
}
\label{tab:genOODvsRealOOD}
\end{table}

\noindent
\textbf{RQ6 Original Images vs.  \rev{Error-inducing} OOD images:}
\revision{In this experiment, we compare \rev{the accuracy of our SUTs (image classifiers) on the error-inducing images 
	generated by \approach versus the} 
	corresponding original images from the dataset. In particular, we examine whether the \rev{SUTs 
	correctly} identify the objects from the non-mutated classes that are present in the image. We then compare 
	the relative performance of each SUT across all non-mutated classes.}

\revision{For non-mutated classes, we found that \rev{\approach's mutation
 impairs SUT accuracy on the original images by up to 24.2\%, on average}.  
\autoref{tab:errorOODAcc} shows that the accuracy of the SUT on \rev{error-inducing} images is broadly lower than \rev{their} accuracy 
	on the corresponding original images. 
Furthermore, we find that the performance of our SUTs on CityScapes is worse than the performance on MSCOCO. We attribute this to the smaller set of classes found in CityScapes (30 as opposed to 183 in MSCOCO). For instance, certain objects (not present in the ground truth) recognised by the SUTs are excluded from the calculation of accuracy in \autoref{tab:errorOODAcc}. Such objects are present with higher frequency in CityScapes due to the relatively smaller set of classes in the ground truth. As such, the accuracy of our SUTs is lower for CityScapes as compared to MSCOCO.
}

\revision{
	We also find instances 
	where the accuracy of the error-inducing images is similar to the accuracy on the original images. For instance, 
	we see that AWS actually performs slightly better (1.4\%) on the error-inducing images generated by \approach 
	on the MSCOCO dataset. The difference in accuracy can be attributed to the fact that these images are 
	error-inducing and as such inherently contain classes that the SUTs struggle to classify. 
In addition, our mutation operators can introduce artifacts that affect the realism of the generated images.	
In conjunction with these artifacts, our insertion operator might also inadvertently cause occlusion that prevent SUTs from performing to their fullest abilities. 
However, considering sufficiently large samples of original images, we note that these 
	\rev{artifacts and occlusions} ought to affect the different classes to a similar extent. As such, we should expect 
	the performance of any particular class to drop by a similar amount when subject to these mutations.}

\begin{result}
\rev{
On average, 
\approach's mutation reduces the accuracy of the SUT on the OOD generated images versus the original images by up to 24.2\% for non-mutated classes.
}
\end{result}

\begin{table}[]
\centering
\caption{Accuracy of our SUTs on non-mutated classes on error-inducing OOD images generated by \approach and the corresponding original images from the dataset.}
\resizebox{\linewidth}{!}{%
\begin{tabular}{@{}ccccccc@{}}
\toprule
\multicolumn{1}{l}{} & \multicolumn{3}{c}{\textbf{MSCOCO}} & \multicolumn{3}{c}{\textbf{CityScapes}} \\ \midrule
\multicolumn{1}{l}{} & \textbf{\begin{tabular}{@{}c@{}}Original \\Images\end{tabular}} & \textbf{\begin{tabular}{@{}c@{}}OOD \\Images\end{tabular}} & \textbf{\begin{tabular}{@{}c@{}}Relative \\Acc. (\%)\end{tabular}} & \textbf{\begin{tabular}{@{}c@{}}Original \\Images\end{tabular}} & \textbf{\begin{tabular}{@{}c@{}}OOD \\Images\end{tabular}} & \textbf{\begin{tabular}{@{}c@{}}Relative \\Acc. (\%)\end{tabular}} \\ \midrule
\textbf{GCP} & 0.469 & 0.434 & 92.5 & 0.345 & 0.262 & 75.8 \\ \midrule
\textbf{MS} & 0.344 & 0.285 & 82.9 & 0.292 & 0.237 & 81.1 \\ \midrule
\textbf{AWS} & 0.649 & 0.658 & 101.4 & 0.656 & 0.463 & 70.5 \\ \midrule
\textbf{Average} & 0.487 & 0.459 & 92.3 & 0.431 & 0.321 & 75.8 \\ \bottomrule
\end{tabular}
}
\label{tab:errorOODAcc}
\end{table}

\section{Limitations and Threats to Validity}
\label{sec:threats}

\noindent
\textbf{Internal Validity:} The main threat to internal validity %
is whether our implementation indeed performs OOD-based test generation. %
We mitigate this threat by conducting typical software quality controls such as testing and code review. For instance, 
we ran several tests to ensure our implementation produced the expected outcome for each mutation, dataset and 
subject program. We also manually inspected random samples of generated images and compare them to the original 
image to ensure our mutation operations are indeed OOD and related to class-level fairness. %
Finally, we conducted a user study %
to examine the semantic validity  of OOD images ({\bf RQ4}). 

\noindent
\textbf{Construct Validity:} This relates to the metrics and measures employed in our experimental analysis. 
We mitigate 
this by employing standard measures of test generation effectiveness such as the number/rate of generated inputs, 
error-inducing inputs and fairness errors (or violations). Such measures are employed in the literature to evaluate 
fairness testing and test generation methods~\cite{chen2022fairness,hort2022bia,soremekun2022software}. 
\revision{Additionally, our ground truth (GT) oracle implicitly relies on the accuracy of the labels in the dataset. These labels are typically labelled by humans. We mitigate this threat by having an alternative metamorphic (MT) oracle. We find that there is a substantial (67\%) overlap between the unfair classes found by the two oracles.}

\noindent
\textbf{External Validity:} %
We acknowledge that \approach may not generalize to all image datasets and image classifiers. However, we have 
evaluated our approach with well-known, commonly used  datasets~\cite{soremekun2022software} 
(\textit{see} \autoref{tab:eval-datasets}). In addition, our subjects are off-the-shelf, mature, commercial image classifiers 
provided by software companies such as Google, Amazon and Microsoft (\textit{see} \autoref{tab:subj-programs}). 

\noindent
\textbf{\reviseCheck{Realism of Mutation Operators:}}
\reviseCheck{
The images generated by \approach can contain elements that are inconsistent with the unmodified portions of the image. For instance, the objects introduced by the insertion operator could have been exposed to different lighting conditions than the objects already in the image. The lack of appropriate shadow detail could conceivably lead to more errors. 
\revision{We also note that in some cases \approach could conceivably attempt to remove a large portion of the original image due to the objects belonging to the mutating class making up most of the image. In such cases, the images generated by \approach could be more unrealistic.}
 We control for this by evaluating against an in-distribution baseline in RQ2. We also evaluate \approach against OOD images present in the dataset in RQ5 and find that our mutation does not adversely affect the detection accuracy of the SUTs. This allows us to compare whether the OOD nature of the images is indeed causing the errors detected. Similarly, both the deletion and rotation can introduce artifacts that affect the realism of the generated images. %
 \revision{Finally, our evaluation in RQ4 both subjectively (via user study) and objectively (via CLIP-IQA~\cite{wang2023exploring})} shows that our images are largely realistic, but we acknowledge that the mutation operators could benefit from better techniques that generate more realistic images.
}

\section{Related Work}
\label{sec:related-work}

\noindent
\textbf{Fairness Test Generation:}
Recent surveys~\cite{soremekun2022software,hort2022bia,chen2022fairness} on software fairness show that researchers employ different 
software analysis and model analysis methods to expose bias in ML systems. 
On one hand, white box fairness testing approaches employ ML techniques (e.g., gradient computation, and clustering) to generate discriminatory 
test cases (e.g., ADF~\cite{zhang2020white, zhang2021automatic} and EIDIG~\cite{zhang2021efficient}). 
On the other hand, black-box approaches leverage the input space and search algorithms to generate discriminatory inputs, e.g., using schemas, grammar, mutation or search algorithms to drive fairness test generation~\cite{udeshi2018automated, yang2021biasrv, soremekun2022astraea, sun2020automatic}. 
Grey-box fairness testing approaches~\cite{tizpaz2022fairness} employ both input space exploration and model analysis for test 
generation.  %
Besides, some methods employ program analysis techniques, e.g., symbolic execution~\cite{aggarwal2019black} 
and combinatorial testing~\cite{morales2021coverage} to expose bias in  ML systems. 
Likewise, we propose a black-box fairness test generation approach. Albeit, unlike prior works, 
we focus on fairness test generation for image recognition systems using distribution-aware and semantic-preserving mutations. %

\noindent
\textbf{OOD Sampling, Distribution-aware \& OOD Testing:} %
Empirical studies on OOD testing have shown that it is important for test generation and revealing faults in ML systems. 
For instance, Berend et al.~\cite{berend2020cats} found that data distribution awareness in both testing and enhancement phases 
outperforms distribution unaware retraining. 
Likewise, Zhou et al.~\cite{zhou2020empirical} showed that OOD-aware detection modules have better performance and are more 
robust against random noises. Similar to these works, we show that OOD testing is important for automatically revealing faults in ML 
systems. %
Berend et al.~\cite{berend2021distribution} 
proposed a distribution aware robustness testing tool to generate unseen test cases for ML task and recommends that ML testing tools 
should be aware of distribution. Besides, Huang et al.~\cite{huang2022hierarchical}  proposed a distribution-aware robustness testing 
approach for detecting adversarial examples using the input distribution and the perceptual quality of inputs. %
This work, unlike \approach, focused on adversarial testing of ML, and not fairness testing. 

\reviseThree{
\recheck{Besides,  Ackerman et al.~\cite{data-slice} proposed an approach to find  explainable data slices where a model underperforms.  In contrast 
to this work, the objective of \approach is to generate test inputs (images) for finding fairness errors. } Additionally, the work by Ackerman et al.~\cite{data-slice} 
neither generates tests beyond the dataset nor does it focus on fairness.  Finally, Vernekar et al.~\cite{ood-sample-gan} proposes an 
approach to generate OOD samples with the objective of improving the accuracy of classifiers on MNIST and Fashion-MNIST datasets. 
Our objective is orthogonal to this work. Specifically, we aim to generate tests that uncover fairness violations in commercial image 
recognition software. As such, \approach proposes an efficient algorithm to generate class-level OOD images based on the occurrences 
and orientations of the class in an existing dataset. Moreover, the work proposed by Vernekar et al.~\cite{ood-sample-gan} does not 
target fairness and only involves simple background and object pixel manipulations. In contrast, we employ insertion, deletion and 
rotation of arbitrary objects in an image, as such is crucial to detect the \recheck{statistical disparity among} class-level accuracy (i.e., fairness).}

\noindent
\textbf{\reviseCheck{Testing of Image Recognition Systems:}}
\reviseCheck{
Researchers have leveraged traditional software testing approaches to test image recognition systems in recent years. For instance, MetaOD leverages an insertion operation to surface errors in object detection systems~\cite{metamorphic-image-ase2020}. Studies have also demonstrated the benefits of applying image modification techniques to computer vision systems~\cite{semantic-image-fuzzing-icse2022, wotawa2021framework, image_captioning_issta2022}. Similarly, we propose an automated testing system focused on object recognition. However, we seek to uncover fairness errors as opposed to the functional errors uncovered by previous works.
}

\noindent
\textbf{Fairness Analysis of Image Recognition Systems:}
Several works have studied and analysed bias in image recognition systems~\cite{yu2020fair, wang2020towards, kim2019learning, brandao2019age, de2019does, denton2019detecting, wang2019balanced}. For instance, 
DeepFAIT~\cite{zhang2021fairness} is a white-box fairness testing approach that requires access to the software at hand, 
which is not applicable for real-world commercial software systems such as our subject programs. Similar to our work, 
Guehairia et al.~\cite{guehairia2022facial} also proposed an OOD detection approach for fairness analysis of facial 
recognition systems. The focus of this work is to enable fair dataset curation and data augmentation rather than test generation. 
In addition, DeepInspect~\cite{tian2020testing} %
exposes  class-level confusion and bias errors  in image classifiers. 
Unlike \approach, DeepInspect is a white-box approach that does not generate a new test 
suite for image classifiers. Instead, it analyzes  image classifiers using {\em only} 
an existing dataset to determine class-level violations.

\section{Conclusion}
\label{sec:conclusion}

In this paper, we propose \approach, a systematic approach to discover class-level fairness violations in image 
classification tasks. The crux of \approach is OOD test generation, which is synergistically combined with 
semantic preserving mutation operations. We show that such an approach is highly effective in revealing 
class-level fairness violations (at least 21\% of generated tests reveal fairness errors) and it significantly outperforms test 
generation within the distribution (2.3x more effective). Additionally, we show that our generated tests 
(OOD images) are 80\% as realistic as real world images. Even though we apply our approach for image classification 
tasks, we believe that our approach
is generally applicable for 
validating multi-label object classification tasks in other domains. We hope that our open source OOD testing 
platform unfolds new opportunities for simple, yet effective class-level fairness testing for a variety of ML 
software systems.

\section{Data Availability}
\label{sec:data}

We will make the experimental data and source code publicly available on acceptance.
In line with that, we provide \approach and our
experimental data for easy reproducibility, reuse and scrutiny:
\begin{center}
	\url{https://github.com/sparkssss/DistroFair}
\end{center}

\IEEEpeerreviewmaketitle

\balance

\bibliographystyle{plainurl}

\bibliography{ms}

\begin{thebibliography}{10}

\bibitem{data-slice}
Samuel Ackerman, Orna Raz, and Marcel Zalmanovici.
\newblock Freaai: Automated extraction of data slices to test machine learning
  models.
\newblock In {\em Engineering Dependable and Secure Machine Learning Systems:
  Third International Workshop, EDSMLS 2020, New York City, NY, USA, February
  7, 2020, Revised Selected Papers}, pages 67--83. Springer, 2020.

\bibitem{aggarwal2019black}
Aniya Aggarwal, Pranay Lohia, Seema Nagar, Kuntal Dey, and Diptikalyan Saha.
\newblock Black box fairness testing of machine learning models.
\newblock In {\em Proceedings of the 2019 27th ACM Joint Meeting on European
  Software Engineering Conference and Symposium on the Foundations of Software
  Engineering}, pages 625--635, 2019.

\bibitem{OOD-engineering}
Aishwarya Agrawal, Dhruv Batra, Devi Parikh, and Aniruddha Kembhavi.
\newblock Don't just assume; look and answer: Overcoming priors for visual
  question answering.
\newblock In {\em 2018 {IEEE} Conference on Computer Vision and Pattern
  Recognition, {CVPR} 2018, Salt Lake City, UT, USA, June 18-22, 2018}, pages
  4971--4980. Computer Vision Foundation / {IEEE} Computer Society, 2018.

\bibitem{aws-vision}
Amazon.
\newblock Amazon rekognition api, 2023.
\newblock URL: \url{https://aws.amazon.com/rekognition/}.

\bibitem{berend2021distribution}
David Berend.
\newblock Distribution awareness for ai system testing.
\newblock In {\em 2021 IEEE/ACM 43rd International Conference on Software
  Engineering: Companion Proceedings (ICSE-Companion)}, pages 96--98. IEEE,
  2021.

\bibitem{berend2020cats}
David Berend, Xiaofei Xie, Lei Ma, Lingjun Zhou, Yang Liu, Chi Xu, and Jianjun
  Zhao.
\newblock Cats are not fish: Deep learning testing calls for
  out-of-distribution awareness.
\newblock In {\em Proceedings of the 35th IEEE/ACM International Conference on
  Automated Software Engineering}, pages 1041--1052, 2020.

\bibitem{bertalmio2000image}
Marcelo Bertalmio, Guillermo Sapiro, Vincent Caselles, and Coloma Ballester.
\newblock Image inpainting.
\newblock In {\em Proceedings of the 27th annual conference on Computer
  graphics and interactive techniques}, pages 417--424, 2000.

\bibitem{brandao2019age}
Martim Brandao.
\newblock Age and gender bias in pedestrian detection algorithms.
\newblock {\em arXiv preprint arXiv:1906.10490}, 2019.

\bibitem{chen2022fairness}
Zhenpeng Chen, Jie~M Zhang, Max Hort, Federica Sarro, and Mark Harman.
\newblock Fairness testing: A comprehensive survey and analysis of trends.
\newblock {\em arXiv preprint arXiv:2207.10223}, 2022.

\bibitem{Cordts2016Cityscapes}
Marius Cordts, Mohamed Omran, Sebastian Ramos, Timo Rehfeld, Markus Enzweiler,
  Rodrigo Benenson, Uwe Franke, Stefan Roth, and Bernt Schiele.
\newblock The cityscapes dataset for semantic urban scene understanding.
\newblock In {\em Proc. of the IEEE Conference on Computer Vision and Pattern
  Recognition (CVPR)}, 2016.

\bibitem{de2019does}
Terrance De~Vries, Ishan Misra, Changhan Wang, and Laurens Van~der Maaten.
\newblock Does object recognition work for everyone?
\newblock In {\em Proceedings of the IEEE/CVF Conference on Computer Vision and
  Pattern Recognition Workshops}, pages 52--59, 2019.

\bibitem{denton2019detecting}
Emily Denton, Ben Hutchinson, Margaret Mitchell, and Timnit Gebru.
\newblock Detecting bias with generative counterfactual face attribute
  augmentation.
\newblock 2019.

\bibitem{fan2014automated}
Shaojing Fan, Tian-Tsong Ng, Jonathan~S Herberg, Bryan~L Koenig, Cheston Y-C
  Tan, and Rangding Wang.
\newblock An automated estimator of image visual realism based on human
  cognition.
\newblock In {\em Proceedings of the IEEE Conference on Computer Vision and
  Pattern Recognition}, pages 4201--4208, 2014.

\bibitem{gadde2021detail}
Raghudeep Gadde, Qianli Feng, and Aleix~M Martinez.
\newblock Detail me more: Improving gan's photo-realism of complex scenes.
\newblock In {\em Proceedings of the IEEE/CVF International Conference on
  Computer Vision}, pages 13950--13959, 2021.

\bibitem{gal2016uncertainty}
Yarin Gal.
\newblock {\em Uncertainty in Deep Learning}.
\newblock PhD thesis, University of Cambridge, 2016.

\bibitem{google-vision}
Google.
\newblock Google cloud vision api, 2023.
\newblock URL: \url{https://cloud.google.com/vision}.

\bibitem{OOD-model}
Yash Goyal, Tejas Khot, Douglas Summers{-}Stay, Dhruv Batra, and Devi Parikh.
\newblock Making the {V} in {VQA} matter: Elevating the role of image
  understanding in visual question answering.
\newblock In {\em 2017 {IEEE} Conference on Computer Vision and Pattern
  Recognition, {CVPR} 2017, Honolulu, HI, USA, July 21-26, 2017}, pages
  6325--6334. {IEEE} Computer Society, 2017.

\bibitem{guehairia2022facial}
O~Guehairia, F~Dornaika, A~Ouamane, and Abdelmalik Taleb-Ahmed.
\newblock Facial age estimation using tensor based subspace learning and deep
  random forests.
\newblock {\em Information Sciences}, 2022.

\bibitem{He_2016_CVPR}
Kaiming He, Xiangyu Zhang, Shaoqing Ren, and Jian Sun.
\newblock Deep residual learning for image recognition.
\newblock In {\em Proceedings of the IEEE Conference on Computer Vision and
  Pattern Recognition (CVPR)}, June 2016.

\bibitem{hort2022bia}
Max Hort, Zhenpeng Chen, Jie~M Zhang, Federica Sarro, and Mark Harman.
\newblock Bias mitigation for machine learning classifiers: A comprehensive
  survey.
\newblock {\em arXiv preprint arXiv:2207.07068}, 2022.

\bibitem{huang2022hierarchical}
Wei Huang, Xingyu Zhao, Alec Banks, Victoria Cox, and Xiaowei Huang.
\newblock Hierarchical distribution-aware testing of deep learning.
\newblock {\em arXiv preprint arXiv:2205.08589}, 2022.

\bibitem{jiang2022improving}
Chiyu~Max Jiang, Mahyar Najibi, Charles~R Qi, Yin Zhou, and Dragomir Anguelov.
\newblock Improving the intra-class long-tail in 3d detection via rare example
  mining.
\newblock In {\em European Conference on Computer Vision}, pages 158--175.
  Springer, 2022.

\bibitem{jin2022pseudo}
Xin Jin, Hao Lou, Heng Huang, Xinning Li, Xiaodong Li, Shuai Cui, Xiaokun
  Zhang, and Xiqiao Li.
\newblock Pseudo-labeling and meta reweighting learning for image aesthetic
  quality assessment.
\newblock {\em IEEE Transactions on Intelligent Transportation Systems},
  23(12):25226--25235, 2022.

\bibitem{kastryulin2022piq}
Sergey Kastryulin, Jamil Zakirov, Denis Prokopenko, and Dmitry~V. Dylov.
\newblock Pytorch image quality: Metrics for image quality assessment, 2022.
\newblock URL: \url{https://arxiv.org/abs/2208.14818}, \href
  {https://doi.org/10.48550/ARXIV.2208.14818}
  {\path{doi:10.48550/ARXIV.2208.14818}}.

\bibitem{kim2019learning}
Byungju Kim, Hyunwoo Kim, Kyungsu Kim, Sungjin Kim, and Junmo Kim.
\newblock Learning not to learn: Training deep neural networks with biased
  data.
\newblock In {\em Proceedings of the IEEE/CVF Conference on Computer Vision and
  Pattern Recognition}, pages 9012--9020, 2019.

\bibitem{Kirillov_2019_CVPR}
Alexander Kirillov, Kaiming He, Ross Girshick, Carsten Rother, and Piotr
  Dollar.
\newblock Panoptic segmentation.
\newblock In {\em Proceedings of the IEEE/CVF Conference on Computer Vision and
  Pattern Recognition (CVPR)}, June 2019.

\bibitem{li2021method}
Cui-jin Li, Zhong Qu, Sheng-ye Wang, and Ling Liu.
\newblock A method of cross-layer fusion multi-object detection and recognition
  based on improved faster r-cnn model in complex traffic environment.
\newblock {\em Pattern Recognition Letters}, 145:127--134, 2021.

\bibitem{lin2018st}
Chen-Hsuan Lin, Ersin Yumer, Oliver Wang, Eli Shechtman, and Simon Lucey.
\newblock St-gan: Spatial transformer generative adversarial networks for image
  compositing.
\newblock In {\em Proceedings of the IEEE conference on computer vision and
  pattern recognition}, pages 9455--9464, 2018.

\bibitem{lin2014microsoft}
Tsung-Yi Lin, Michael Maire, Serge Belongie, James Hays, Pietro Perona, Deva
  Ramanan, Piotr Doll{\'a}r, and C~Lawrence Zitnick.
\newblock Microsoft coco: Common objects in context.
\newblock In {\em European conference on computer vision}, pages 740--755.
  Springer, 2014.

\bibitem{ma2017end}
Kede Ma, Wentao Liu, Kai Zhang, Zhengfang Duanmu, Zhou Wang, and Wangmeng Zuo.
\newblock End-to-end blind image quality assessment using deep neural networks.
\newblock {\em IEEE Transactions on Image Processing}, 27(3):1202--1213, 2017.

\bibitem{kmeans}
J~MacQueen.
\newblock Classification and analysis of multivariate observations.
\newblock In {\em 5th Berkeley Symp. Math. Statist. Probability}, pages
  281--297, 1967.

\bibitem{michaelis2019benchmarking}
Claudio Michaelis, Benjamin Mitzkus, Robert Geirhos, Evgenia Rusak, Oliver
  Bringmann, Alexander~S Ecker, Matthias Bethge, and Wieland Brendel.
\newblock Benchmarking robustness in object detection: Autonomous driving when
  winter is coming.
\newblock {\em arXiv preprint arXiv:1907.07484}, 2019.

\bibitem{ms-vision}
Microsoft.
\newblock Azure computer vision api, 2023.
\newblock URL:
  \url{https://azure.microsoft.com/en-us/services/cognitive-services/computer-vision/}.

\bibitem{morales2021coverage}
Daniel~Perez Morales, Takashi Kitamura, and Shingo Takada.
\newblock Coverage-guided fairness testing.
\newblock In {\em International Conference on Intelligence Science}, pages
  183--199. Springer, 2021.

\bibitem{mturk}
Amazon MTurk.
\newblock Amazon mechanical turk, 2023.
\newblock URL: \url{https://www.mturk.com/}.

\bibitem{deepexplore}
Kexin Pei, Yinzhi Cao, Junfeng Yang, and Suman Jana.
\newblock Deepxplore: Automated whitebox testing of deep learning systems.
\newblock In {\em Proceedings of the 26th Symposium on Operating Systems
  Principles, Shanghai, China, October 28-31, 2017}, pages 1--18. {ACM}, 2017.

\bibitem{racist-camera}
Adam Rose.
\newblock Are face-detection cameras racist?
\newblock
  \url{http://content.time.com/time/business/article/0,8599,1954643,00.html},
  2010.

\bibitem{scikit}
scikit learn:.
\newblock scikit-learn: Machine learning in python, 2023.
\newblock URL: \url{https://scikit-learn.org/stable/}.

\bibitem{soremekun2022software}
Ezekiel Soremekun, Mike Papadakis, Maxime Cordy, and Yves~Le Traon.
\newblock Software fairness: An analysis and survey.
\newblock {\em arXiv preprint arXiv:2205.08809}, 2022.

\bibitem{soremekun2022astraea}
Ezekiel Soremekun, Sakshi~Sunil Udeshi, and Sudipta Chattopadhyay.
\newblock Astraea: Grammar-based fairness testing.
\newblock {\em IEEE Transactions on Software Engineering}, 2022.

\bibitem{sun2020automatic}
Zeyu Sun, Jie~M Zhang, Mark Harman, Mike Papadakis, and Lu~Zhang.
\newblock Automatic testing and improvement of machine translation.
\newblock In {\em Proceedings of the ACM/IEEE 42nd International Conference on
  Software Engineering}, pages 974--985, 2020.

\bibitem{suvorov2022resolution}
Roman Suvorov, Elizaveta Logacheva, Anton Mashikhin, Anastasia Remizova,
  Arsenii Ashukha, Aleksei Silvestrov, Naejin Kong, Harshith Goka, Kiwoong
  Park, and Victor Lempitsky.
\newblock Resolution-robust large mask inpainting with fourier convolutions.
\newblock In {\em Proceedings of the IEEE/CVF Winter Conference on Applications
  of Computer Vision}, pages 2149--2159, 2022.

\bibitem{vqa-cp}
Damien Teney, Ehsan Abbasnejad, Kushal Kafle, Robik Shrestha, Christopher
  Kanan, and Anton van~den Hengel.
\newblock On the value of out-of-distribution testing: An example of goodhart's
  law.
\newblock In {\em Advances in Neural Information Processing Systems 33: Annual
  Conference on Neural Information Processing Systems 2020}, 2020.

\bibitem{deeptest}
Yuchi Tian, Kexin Pei, Suman Jana, and Baishakhi Ray.
\newblock Deeptest: automated testing of deep-neural-network-driven autonomous
  cars.
\newblock In Michel Chaudron, Ivica Crnkovic, Marsha Chechik, and Mark Harman,
  editors, {\em Proceedings of the 40th International Conference on Software
  Engineering, {ICSE} 2018, Gothenburg, Sweden, May 27 - June 03, 2018}, pages
  303--314. {ACM}, 2018.

\bibitem{tian2020testing}
Yuchi Tian, Ziyuan Zhong, Vicente Ordonez, Gail~E. Kaiser, and Baishakhi Ray.
\newblock Testing {DNN} image classifiers for confusion {\&} bias errors.
\newblock In {\em {ICSE} '20: 42nd International Conference on Software
  Engineering, Seoul, South Korea, 27 June - 19 July, 2020}, pages 1122--1134.
  {ACM}, 2020.

\bibitem{tizpaz2022fairness}
Saeid Tizpaz-Niari, Ashish Kumar, Gang Tan, and Ashutosh Trivedi.
\newblock Fairness-aware configuration of machine learning libraries.
\newblock In {\em 2022 IEEE/ACM 44th International Conference on Software
  Engineering (ICSE)}. IEEE, 2022.

\bibitem{treml2016speeding}
Michael Treml, Jos{\'e} Arjona-Medina, Thomas Unterthiner, Rupesh Durgesh,
  Felix Friedmann, Peter Schuberth, Andreas Mayr, Martin Heusel, Markus
  Hofmarcher, Michael Widrich, et~al.
\newblock Speeding up semantic segmentation for autonomous driving.
\newblock 2016.

\bibitem{udeshi2018automated}
Sakshi Udeshi, Pryanshu Arora, and Sudipta Chattopadhyay.
\newblock Automated directed fairness testing.
\newblock In Marianne Huchard, Christian K{\"{a}}stner, and Gordon Fraser,
  editors, {\em Proceedings of the 33rd {ACM/IEEE} International Conference on
  Automated Software Engineering, {ASE} 2018, Montpellier, France, September
  3-7, 2018}, pages 98--108. {ACM}, 2018.

\bibitem{verma2018fairness}
Sahil Verma and Julia Rubin.
\newblock Fairness definitions explained.
\newblock In {\em 2018 ieee/acm international workshop on software fairness
  (fairware)}, pages 1--7. IEEE, 2018.

\bibitem{ood-sample-gan}
Sachin Vernekar, Ashish Gaurav, Vahdat Abdelzad, Taylor Denouden, Rick Salay,
  and Krzysztof Czarnecki.
\newblock Out-of-distribution detection in classifiers via generation.
\newblock 2019.

\bibitem{wang2023exploring}
Jianyi Wang, Kelvin~CK Chan, and Chen~Change Loy.
\newblock Exploring clip for assessing the look and feel of images.
\newblock In {\em Proceedings of the AAAI Conference on Artificial
  Intelligence}, volume~37, pages 2555--2563, 2023.

\bibitem{metamorphic-image-ase2020}
Shuai Wang and Zhendong Su.
\newblock Metamorphic object insertion for testing object detection systems.
\newblock In {\em 35th {IEEE/ACM} International Conference on Automated
  Software Engineering, {ASE} 2020, Melbourne, Australia, September 21-25,
  2020}, pages 1053--1065. {IEEE}, 2020.

\bibitem{wang2019balanced}
Tianlu Wang, Jieyu Zhao, Mark Yatskar, Kai-Wei Chang, and Vicente Ordonez.
\newblock Balanced datasets are not enough: Estimating and mitigating gender
  bias in deep image representations.
\newblock In {\em Proceedings of the IEEE/CVF International Conference on
  Computer Vision}, pages 5310--5319, 2019.

\bibitem{wang2020towards}
Zeyu Wang, Klint Qinami, Ioannis~Christos Karakozis, Kyle Genova, Prem Nair,
  Kenji Hata, and Olga Russakovsky.
\newblock Towards fairness in visual recognition: Effective strategies for bias
  mitigation.
\newblock In {\em Proceedings of the IEEE/CVF conference on computer vision and
  pattern recognition}, pages 8919--8928, 2020.

\bibitem{semantic-image-fuzzing-icse2022}
Trey Woodlief, Sebastian~G. Elbaum, and Kevin Sullivan.
\newblock Semantic image fuzzing of {AI} perception systems.
\newblock In {\em 44th {IEEE/ACM} 44th International Conference on Software
  Engineering, {ICSE} 2022, Pittsburgh, PA, USA, May 25-27, 2022}, pages
  1958--1969. {ACM}, 2022.

\bibitem{wotawa2021framework}
Franz Wotawa, Lorenz Klampfl, and Ledio Jahaj.
\newblock A framework for the automation of testing computer vision systems.
\newblock In {\em 2021 IEEE/ACM International Conference on Automation of
  Software Test (AST)}, pages 121--124. IEEE, 2021.

\bibitem{wu2020multi}
Jian Wu, Victor~S Sheng, Jing Zhang, Hua Li, Tetiana Dadakova, Christine~Leon
  Swisher, Zhiming Cui, and Pengpeng Zhao.
\newblock Multi-label active learning algorithms for image classification:
  Overview and future promise.
\newblock {\em ACM Computing Surveys (CSUR)}, 53(2):1--35, 2020.

\bibitem{wu2019detectron2}
Yuxin Wu, Alexander Kirillov, Francisco Massa, Wan-Yen Lo, and Ross Girshick.
\newblock Detectron2.
\newblock \url{https://github.com/facebookresearch/detectron2}, 2019.

\bibitem{xiong2019upsnet}
Yuwen Xiong, Renjie Liao, Hengshuang Zhao, Rui Hu, Min Bai, Ersin Yumer, and
  Raquel Urtasun.
\newblock Upsnet: A unified panoptic segmentation network.
\newblock In {\em Proceedings of the IEEE/CVF Conference on Computer Vision and
  Pattern Recognition}, pages 8818--8826, 2019.

\bibitem{yang2021biasrv}
Zhou Yang, Muhammad~Hilmi Asyrofi, and David Lo.
\newblock Biasrv: Uncovering biased sentiment predictions at runtime.
\newblock In {\em Proceedings of the 29th ACM Joint Meeting on European
  Software Engineering Conference and Symposium on the Foundations of Software
  Engineering}, pages 1540--1544, 2021.

\bibitem{ood-bench}
Nanyang Ye, Kaican Li, Haoyue Bai, Runpeng Yu, Lanqing Hong, Fengwei Zhou,
  Zhenguo Li, and Jun Zhu.
\newblock Ood-bench: Quantifying and understanding two dimensions of
  out-of-distribution generalization.
\newblock In {\em {CVPR}}, pages 7937--7948. {IEEE}, 2022.

\bibitem{image_captioning_issta2022}
Boxi Yu, Zhiqing Zhong, Xinran Qin, Jiayi Yao, Yuancheng Wang, and Pinjia He.
\newblock Automated testing of image captioning systems.
\newblock In {\em {ISSTA}}, pages 467--479. {ACM}, 2022.

\bibitem{yu2020fair}
Jun Yu, Xinlong Hao, Haonian Xie, and Ye~Yu.
\newblock Fair face recognition using data balancing, enhancement and fusion.
\newblock In {\em European Conference on Computer Vision}, pages 492--505.
  Springer, 2020.

\bibitem{zhang2021efficient}
Lingfeng Zhang, Yueling Zhang, and Min Zhang.
\newblock Efficient white-box fairness testing through gradient search.
\newblock In {\em Proceedings of the 30th ACM SIGSOFT International Symposium
  on Software Testing and Analysis}, pages 103--114, 2021.

\bibitem{zhang2020white}
Peixin Zhang, Jingyi Wang, Jun Sun, Guoliang Dong, Xinyu Wang, Xingen Wang,
  Jin~Song Dong, and Ting Dai.
\newblock White-box fairness testing through adversarial sampling.
\newblock In {\em {ICSE} '20: 42nd International Conference on Software
  Engineering, Seoul, South Korea, 27 June - 19 July, 2020}, pages 949--960.
  {ACM}, 2020.

\bibitem{zhang2021fairness}
Peixin Zhang, Jingyi Wang, Jun Sun, and Xinyu Wang.
\newblock Fairness testing of deep image classification with adequacy metrics.
\newblock {\em arXiv preprint arXiv:2111.08856}, 2021.

\bibitem{zhang2021automatic}
Peixin Zhang, Jingyi Wang, Jun Sun, Xinyu Wang, Guoliang Dong, Xingen Wang,
  Ting Dai, and Jin~Song Dong.
\newblock Automatic fairness testing of neural classifiers through adversarial
  sampling.
\newblock {\em IEEE Transactions on Software Engineering}, 2021.

\bibitem{men-like-shopping}
Jieyu Zhao, Tianlu Wang, Mark Yatskar, Vicente Ordonez, and Kai{-}Wei Chang.
\newblock Men also like shopping: Reducing gender bias amplification using
  corpus-level constraints.
\newblock In Martha Palmer, Rebecca Hwa, and Sebastian Riedel, editors, {\em
  Proceedings of the 2017 Conference on Empirical Methods in Natural Language
  Processing, {EMNLP} 2017, Copenhagen, Denmark, September 9-11, 2017}, pages
  2979--2989. Association for Computational Linguistics, 2017.

\bibitem{zhou2020empirical}
Lingjun Zhou, Bing Yu, David Berend, Xiaofei Xie, Xiaohong Li, Jianjun Zhao,
  and Xusheng Liu.
\newblock An empirical study on robustness of dnns with out-of-distribution
  awareness.
\newblock In {\em 2020 27th Asia-Pacific Software Engineering Conference
  (APSEC)}, pages 266--275. IEEE, 2020.

\end{thebibliography}

\end{document}